%% file: sample-journal.tex
\documentclass[acmtog]{acmart}
\usepackage{booktabs} %
\usepackage[ruled]{algorithm2e} %
\usepackage{algorithmic}
\usepackage{xcolor}
\usepackage{multirow}
\usepackage{amsfonts}
\usepackage{dblfloatfix}

\setcopyright{acmcopyright}

\def\1{\mathds{1}}

\def\eg{{e.g.$\;$}}

\hypersetup{draft} 
 
\begin{document}
\title{Manipulating Attributes of Natural Scenes via Hallucination}

\author{Levent Karacan}
\orcid{1234-5678-9012-3456}
\affiliation{%
  \institution{Hacettepe University and Iskenderun Technical University}
  \city{Ankara}
  \country{Turkey}}
\email{karacan@cs.hacettepe.edu.tr}
\author{Zeynep Akata}
\affiliation{%
  \institution{University of T\"{u}bingen}
  \city{Tubingen}
  \country{Germany}
}
\email{zeynep.akata@uni-tuebingen.de}
\author{Aykut Erdem}
\affiliation{%
 \institution{Hacettepe University}
 \city{Ankara}
 \country{Turkey}}
\email{aykut@cs.hacettepe.edu.tr}
\author{Erkut Erdem}
\affiliation{%
  \institution{Hacettepe University}
  \city{Ankara}
  \country{Turkey}
}
\email{erkut@cs.hacettepe.edu.tr}

\renewcommand\shortauthors{Karacan, L., Akata Z., Erdem A., Erdem E.}

\begin{abstract}
In this study, we explore building a two-stage framework for enabling users to directly manipulate high-level attributes of a natural scene. The key to our approach is a deep generative network which can hallucinate images of a scene as if they were taken at a different season (e.g. during winter), weather condition (e.g. in a cloudy day) or time of the day (e.g. at sunset). Once the scene is hallucinated with the given attributes, the corresponding look is then transferred to the input image while preserving the semantic details intact, giving a photo-realistic manipulation result. As the proposed framework hallucinates what the scene will look like, it does not require any reference style image as commonly utilized in most of the appearance or style transfer approaches. Moreover, it allows to simultaneously manipulate a given scene according to a diverse set of transient attributes within a single model, eliminating the need of training multiple networks per each translation task. Our comprehensive set of qualitative and quantitative results demonstrate the effectiveness of our approach against the competing methods.
\end{abstract}


\begin{teaserfigure}
\centering
\includegraphics[width=\linewidth]{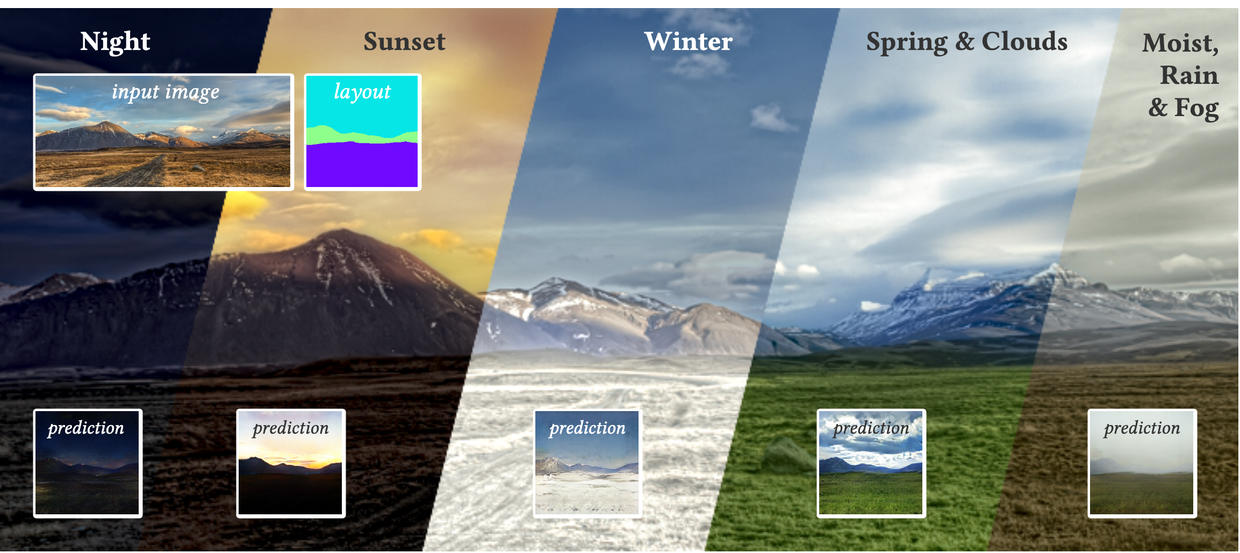}
\caption{Given a natural image, our approach can hallucinate different versions of the same scene in a wide range of conditions, e.g.\emph{night}, \emph{sunset}, \emph{winter}, \emph{spring}, \emph{rain}, \emph{fog} or even a combination of those. First, we utilize a generator network to imagine the scene with respect to its semantic layout and the desired set of attributes. Then, we directly transfer the scene characteristics from the hallucinated output to the input image, without the need for a reference style image.}
\label{fig:teaser}
\end{teaserfigure}

\maketitle

\input{samplebody-journals}

\end{document}

%% file: samplebody-journals.tex
\section{Introduction}
\label{sec:intro}
\begin{quote}
    ``The trees, being partly covered with snow, were outlined indistinctly against the grayish background formed by a cloudy sky, barely whitened by the moon.''\\
    \hspace*{\fill} -- Honore de Balzac (Sarrasine, 1831)
\end{quote}
The visual world we live in constantly changes its appearance depending on time and seasons. For example, at sunset, the sun gets close to the horizon gives the sky a pleasant red tint, with the advent of warm summer, the green tones on the grass leave its place in bright yellowish tones and autumn brings a variety of shades of brown and yellow to the trees. Such visual changes in the nature continues in various forms at almost any moment with the effect of time, weather and season. Such high-level changes are referred to as \emph{transient scene attributes} -- e.g. cloudy, foggy, night, sunset, winter, summer, to name a few~\cite{LRTQH14}. 
 
Recognizing transient attributes of an outdoor image and modifying its content to reflect any changes in these properties were studied in the past, however, current approaches have many constraints which limit their usability and effectiveness in attribute manipulation. In this paper, we present a framework that can hallucinate different versions of a natural scene given its semantic layout and its desired real valued transient attributes. Our model can generate many possible output images from scratch such as the ones in Fig.~\ref{fig:teaser}, which is made possible by learning from data the semantic meaning of each transient attribute and the corresponding local and global transformations.

Image generation is quite a challenging task since it needs to have realistic looking outputs. Visual attribute manipulation can be considered a bit harder as it aims at photorealism as well as results that are semantically consistent with the input image. For example, for predicting the look of a scene at sunset, visual appearances of the sky and the ground undergo changes differently, the sky gets different shades of red while the dominant color of the ground becomes much darker and texture details get lost. Unlike recent image synthesis methods ~\cite{isola2016image,chen2017photographic,wang2018high,qi2018semi}, which explore producing realistic-looking images from semantic layouts, automatically manipulating visual attributes requires modifying the appearance of an input image while preserving object-specific semantic details intact. Some recent style transfer methods achieve this goal to a certain extent but they require a reference style image~\cite{luan2017deep,li2018closed}. 

A simple solution to obtain an automatic style transfer method is to retrieve reference style images with desired attributes from a well-prepared dataset with a rich set of attributes. However, this approach raises new issues that need to be solved such as retrieving images according to desired attributes and semantic layout in an effective way. To overcome these obstacles, we propose to combine neural image synthesis and style transfer approaches to perform visual attribute manipulation. For this purpose, we first devise a conditional image synthesis model that is capable of hallucinating desired attributes on synthetically generated scenes with semantic content similar to the input image and then we resort to a photo style transfer method to transfer the visual look of the hallucinated image to the original input image to produce a resulting image with the desired attributes.

 A rich variety of generative models including Generative Adversarial Networks (GANs)~\cite{GPMXWDOCB14,RMC16,VPT16}, Variational Autoencoders (VAEs)~\cite{KW14,GDGRW15}, and autoregressive models~\cite{MPBS16,oord2016conditional} have been developed to synthesize visually plausible images. Images of higher resolutions, e.g. 256$\times$256, 512$\times$512 or 1024$\times$1024, have also been rendered under improved versions of these frameworks~\cite{RAYLSL16,RAMSSL16,shang2017channel,berthelot2017began,gulrajani2016pixelvae,zhu2017unpaired,chen2017photographic,karras2018style,ProgressiveGANs18}. However, generating diverse, photorealistic and well-controlled images of complex scenes has not yet been fully solved.
For image synthesis, we propose a new conditional GAN based approach to generate a target image which has the same semantic layout with the input image but reflects the desired transient attributes. As shown in Fig.~\ref{fig:teaser}, our approach allows users to manipulate the look of an outdoor scene with respect to a set of transient attributes, owing to a learned manifold of natural images.

To build the aforementioned model, we argue the necessity of better control over the generator network in GAN. We address this issue by conditioning ample concrete information of scene contents to the default GAN framework, deriving our proposed attribute and semantic layout conditioned GAN model. Spatial layout information tells the network where to draw, resulting in clearly-defined object boundaries and  transient scene attributes serve to edit visual properties of a given scene so that we can hallucinate desired attributes for input image in semantically similar generated image.

However, naively importing the side information is insufficient. 
For one, when training the discriminator to distinguish mismatched image-condition pairs, if the condition is randomly sampled, it can easily be too off in describing the image to provide meaningful error derivatives.
To address this issue, we propose to selectively sample mismatched layouts for a given real image, inspired by the practice of hard negative mining~\cite{wang2015unsupervised}.
For another, given the challenging nature of the scene generation problem, adversarial objective alone can struggle to discover a satisfying output distribution. 
Existing works in synthesizing complex images apply the technique of ``feature matching'', or perceptual loss~\cite{chen2017photographic,dosovitskiy2016generating}. 
Here, we also adopt perceptual loss to stabilize and improve adversarial training for more photographic generation but contrasting prior works, our approach employs the layout-invariant features pretrained on segmentation task to ensure consistent layouts between synthesized images and reference images.
For photo style transfer, we use a recent deep learning based approach ~\cite{li2018closed} which transfers visual appearance between same semantic objects in real photos using semantic layout maps.

Our contributions are summarized as follows:
\begin{itemize}
\item We propose a new two-stage visual attribute manipulation framework for changing high-level attributes of a given outdoor image.
\item We develop a conditional GAN variant for generating natural scenes faithful to given semantic layouts and transient attributes. 
\item We build up an outdoor scene dataset annotated with layout and transient attribute labels by combining and annotating images from Transient Attributes~\cite{LRTQH14} and ADE20K~\cite{zhou2016scene}. 
\end{itemize}
Our code and models are publicly available at the project website\footnote{\url{https://hucvl.github.io/attribute_hallucination}}.

\section{Related Work}
\label{sec:related}

\subsection{Image Synthesis}
In the past few years, much progress has been made towards realistic image synthesis; in particularly, different flavors and improved versions of Generative Adversarial Networks (GANs)~\cite{GPMXWDOCB14} have achieved impressive results along this direction. 
\citet{RMC16} were the first to propose a  architecture that can be trained on large scale datasets, which sparked a wave of studies aimed at improving this line of work~\cite{arjovsky2017wasserstein,mao2016multi,salimans2016improved}. \citet{larsen2015autoencoding} integrates adversarial discriminator to VAE framework in an attempt to prevent mode collapsing. Its extension~\cite{shang2017channel} further tackles this issue while improving generation quality and resolution. More recently,~\citet{ProgressiveGANs18} have suggested to use a cascaded set of generators to increase both the photorealism and the resolution of generated images. In the subsequent work, ~\citet{karras2018style} have achieved further improvement in realism and diversity of the generated synthetic images by adopting ideas from style transfer literature~\cite{huang17adain}.

Conditional GANs (CGANs)~\cite{mirza2014conditional} that leverages side information have been widely adopted to generate images under predefined constraints. For example, the recently proposed BigGAN~\cite{brock2018large} generates high quality, high resolution images conditioned on visual classes in ImageNet. \citet{RAYLSL16,RAMSSL16} generate images using natural language descriptions; \citet{antipov2017face} follow similar pipelines to edit a given facial appearance based on age. Pix2pix~\cite{isola2016image} undertakes a different approach to conditional generation that it directly translates one type of image information to another type through an encoder-decoder architecture coupled with adversarial loss; its extension Cycle-GAN~\cite{zhu2017unpaired} conducts similar translation under the assumption that well-aligned image pairs are not available. The design of our image synthesis model resembles CGANs, as opposed to Pix2pix, since those so-called image-to-image translation models are limited in terms of output diversity.  

In the domain of scene generation, the aforementioned Pix2pix \cite{isola2016image} and Cycle-GAN~\cite{zhu2017unpaired} both manage to translate realistic scene images from semantic layouts. However, these models are deterministic, in other words, they can only map one input image to one output image in different domains. Recently, some researchers have proposed multimodal (e.g. BicycleGAN~\cite{zhu2017toward}) or multi-domain (e.g. StarGAN~\cite{StarGAN2018}, MUNIT~\cite{huang2018multimodal}) image-to-image translation models. Both of these approaches have the ability to translate a given input image to multiple possible output images with the use of a single network. However, in BicycleGAN, the users have no control over the generation process other than deciding upon the source and target domains. StarGAN and MUNIT can perform many-to-many translations but these the translations are always carried out between two different modalities. Although these works improve the diversity to a certain degree, they are still limited in the sense that they do not allow to fully control the latent scene characteristics. For instance, these methods can not generate an image with a little bit of sunset and partly cloudy skies from an image taken on a clear day. Our proposed model, on the other hand, allows the users to play with all of the scene attributes with varying degrees of freedom at the same time.

Alternatively, some efforts on image-to-image translation has been made to increase the realism and resolution with multi-scale approaches~\cite{chen2017photographic,qi2018semi,wang2018high,park2019spade}. \citet{wang2018high}'s Pix2pixHD model improves both the resolution and the photorealism of  Pix2pix \cite{isola2016image} by employing multi-scale generator and discriminator networks. Recently, \citet{park2019spade} propose a spatially-adaptive normalization scheme to better preserve semantic information. \citet{qi2018semi} utilize a semi-parametric approach and increase the photorealism of the output images by composing real object segments from a set of training images within an image-to-image synthesis network.~\citet{chen2017photographic} try to achieve realism through a carefully crafted regression objective that maps a single input layout to multiple potential scene outputs. Nonetheless, despite modeling one-to-many relationships, the number of outputs is pre-defined and fixed, which still puts tight constraints on the generation process. As compared to these works, besides taking semantic layout as input, our proposed scene generation network is additionally aware of the transient attributes and the latent random noises characterizing intrinsic properties of the generated outputs. As a consequence, our model is more flexible in generating the same scene content under different conditions such as lighting, weather, and seasons.

From training point of view, a careful selection of ``negative'' pairs, i.e. negative mining, is an essential component in metric learning and ranking~\cite{fu2013survey,shrivastava2016training,li2013bootstrapping}. 
Existing works in CGAN have been using randomly sampled negative image-condition pairs~\cite{RAMSSL16}.
However, such random negative mining strategy has been shown to be inferior to more meticulous negative sampling schemes \cite{bucher2016hard}.
Particularly, the negative pair sampling scheme proposed in our work is inspired by the concept of relevant negative~\cite{li2013bootstrapping}, where the negative examples that are visually similar to positive ones are emphasized more during learning. 

To make the generated images look more similar to the reference images, a common technique is to consider feature matching which is commonly employed through a perceptual loss ~\cite{chen2017photographic,dosovitskiy2016generating,johnson2016perceptual}. 
The perceptual loss in our proposed model distinguishes itself from existing works by matching segmentation invariant features from pre-trained segmentation networks~\cite{zhou2016scene}, leading to diverse generations that comply with the given layouts.
\subsection{Image Editing}
There has been a great effort towards building methods for manipulating visual appearance of a given image. Example-based approaches ~\cite{ pitie2005n, reinhard2001color} use a reference image to transfer color space statistics to input image so that visual appearance of input image looks like the reference image. In contrast to these global color transfer approaches, which require highly consistent reference images with input image, user controllable color transfer techniques were also proposed~\cite{an2010user,dale2009image} to consider spatial layouts of input and reference images. ~\citet{dale2009image} search for some reference images which have similar visual context to input image in a large image dataset to transfer local color from them and then use color transferred image to restore input image. Other local color transfer approaches ~\cite{wu2013content} use the semantic segments to transfer color between regions in reference and input images have same semantic label (e.g. color is transferred from sky region in reference image to sky region in input image).
Some data-driven approaches ~\cite{LRTQH14,shih2013data} leverage the time-lapse video datasets taken for same scene to capture scene variations that occur at different times. ~\citet{shih2013data} aim to give times of day appearances to a given input image, for example converting an input image taken midday to a nice sunset image. They first retrieve the most similar video frame to input scene from dataset as reference frame. Then they find matching patches between reference frame and input image. Lastly, they transfer the variation that occurs between reference frame and desired reference frame which is same scene but taken different time of day to input image. ~\citet{LRTQH14} take a step forward in their work for handling more general variations as transient attributes such as lighting, weather, and seasons.

High-level image editing offers easier and more natural way to casual users to manipulate a given image. Instead of using a reference image either provided by the user or retrieved from a database, learning the image manipulations and high-level attributes for image editing like a human has also attracted researchers.
 ~\citet{berthouzoz2011framework} learn parameters of the basic operations for some manipulations recorded in photoshop as macro to adapt them to new images, for example, applying same skin color correction operation with same parameters for both faces with dark-skinned and light-skinned does not give expected correction.
In contrast to learning image operations for specific editing effects, ~\citet{cheng2014imagespirit} learn the attributes as adjectives and objects as nouns for semantic parsing of an image and further use them for verbal guided image manipulation to indoor images. For example, the verbal command ``\textit{change the floor to wooden}'' modifies the appearance of the floor. Similarly, \citet{LRTQH14} learn to recognize transient attributes for attribute-guided image editing on outdoor images. To modify the look of an input image (e.g. a photo taken in a sunny day), they first locate similar scenes in a dataset they collected and annotated with transient attributes. Then they transfer the desired look (e.g. ``\textit{more winter}'') from the corresponding version of the candidate match images by using an appearance transfer method.~\citet{lee2016automatic} aim to automatically select a subset of style exemplars that will achieve good stylization results by learning a content-to-style mapping between large photo collection and a small style dataset.

Deep learning has fueled a growing literature on employing neural approaches to improve existing image editing problems. Here, we review the studies that are the most relevant to our work. 
~\citet{gatys2016image} have demonstrated how Convolutional Neural Networks (CNNs) effectively encode content and texture separately in feature maps of CNNs trained on large-scale image datasets and have proposed a neural style transfer method to transfer artistic styles from paintings to natural images. Alternatively,~\citet{johnson2016perceptual} train a transformation network to speed up the test time of style transferring together with minimization of perceptual loss between input image and stylized image.~\citet{li2017diversified} consider a deep feed-forward network, which is capable of generating multiple and diverse results within a single network. Recent deep photo style transfer method of~\citet{luan2017deep}, named DPST, aims at providing realism in case of style transfer is made between the real photos. For example, when one wants to make an input photo look like taken in different illumination and weather conditions, a photo-realistic transfer is necessary. It uses semantic labels to prevent semantic inconsistency so that style transfer is carried out between same semantic regions. Recently,~\citet{li2018closed} have proposed another photo style transfer method called FPST, which works significantly faster than DPST. It considers a two-steps process, a stylization step followed by a photorealistic smoothing step, both of each having efficient
closed-form solutions. There are some style transfer networks which are specialized for the editing face images and portraits~\cite{kemelmacher2016transfiguring, liao2017visual, selim2016painting} with new objectives. Nevertheless, these style transfer works limit the users to find an reference photo in which desired style effects exist for desired attributes.

~\citet{yan2016automatic} introduce the first automatic photo adjustment framework based on deep neural networks. They use deep neural network to learn a regressor which transforms the colors for artistic styles especially color adjustment from the image and its stylized version pairs. They define a set of feature descriptors based on pixel, global and semantic levels. In another work,  ~\citet{gharbi2017deep} propose a new neural network architecture to learn image enhancement transformations at low resolution, then they move learned transformations to higher resolution in bilateral space in an edge-preserving manner.

Lastly, building upon conditional GAN model, some image completion works have been proposed to predict missing regions providing global and local context information with multiple discriminator networks~\cite{iizuka2017globally,li2017generative}.

\begin{figure*}[t]
\centering
\includegraphics[width=\linewidth]{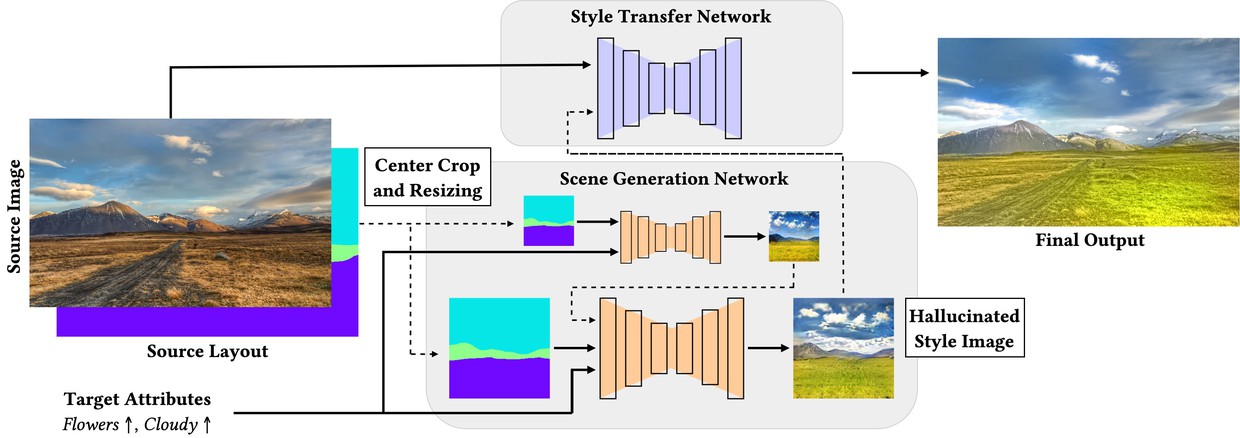}
\vspace{-2mm}
\caption{Overview of the proposed attribute manipulation framework. Given an input image and its semantic layout, we first resize and center-crop the layout to $512 \times 512$ pixels and feed it to our scene generation network. After obtaining the scene synthesized according to the target transient attributes, we transfer the look of the hallucinated style back to the original input image.}
\vspace{-2mm}
\label{fig:system}
\end{figure*}

\section{ALS18K Dataset}
\label{sec:data}
To train our model, we curate a new dataset by selecting and annotating images from two popular scene datasets, namely ADE20K~\cite{zhou2016scene} and Transient Attributes~\cite{LRTQH14}, for the reasons which will become clear shortly. 

ADE20K~\cite{zhou2016scene} includes $22,210$ images from a diverse set of indoor and outdoor scenes which are densely annotated with object and stuff instances from $150$ classes. However, it does not include any information about transient attributes. Transient Attributes \cite{LRTQH14} 
contains $8,571$ outdoor scene images captured by $101$ webcams in which the images of the same scene can exhibit high variance in appearance due to variations in atmospheric conditions caused by weather, time of day, season. The images in this dataset are annotated with 40 transient scene attributes, \eg{sunrise/sunset, cloudy, foggy, autumn, winter}, but this time it lacks semantic layout labels. 

To establish a richly annotated, large-scale dataset of outdoor images with both transient attribute and layout labels, we further operate on these two datasets as follows. First, from ADE20K, we manually pick the 9,201 images corresponding to outdoor scenes, which contain nature and urban scenery pictures. For these images, we need to obtain transient attribute annotations. To do so, we conduct initial attribute predictions using the pretrained model from~\cite{BZGWJ16} and then manually verify the predictions. From Transient Attributes, we select all the 8,571 images. To get the layouts, we first run the semantic segmentation model by~\citet{zhao2017pspnet}, the winner of the MIT Scene Parsing Challenge 2016, and assuming that each webcam image of the same scene has the same semantic layout, we manually select the best semantic layout prediction for each scene and use those predictions as the ground truth layout for the related images.

In total, we collect 17,772 outdoor images (9,201 from ADE20K +  8,571 from Transient Attributes), with 150 semantic categories and 40 transient attributes. Following the train-val split from ADE20K, 8,363 out of the 9,201 images are assigned to the training set, the other 838 testing; for the Transient Attributes dataset, 500 randomly selected images are held out for testing. In total, we have 16,434 training examples and 1,338 testing images. More samples of our annotations are presented in the Supplementary Material. Lastly, we resize the height of all images to 512 pixels and apply center-cropping to obtain $512\times 512$ images.

\section{Attribute Manipulation Framework}
~\label{systemoverview}
Our framework provides an easy and high-level editing system to manipulate transient attributes of outdoor scenes (see Fig.~\ref{fig:system}). The key component of our framework is a scene generation network that is conditioned on semantic layout and continuous-valued vector of transient attributes. This network allows us to generate synthetic scenes consistent with the semantic layout of the input image and having the desired transient attributes. One can play with $40$ different transient attributes by increasing or decreasing values of certain dimensions. Note that, at this stage, the semantic layout of the input image should also be fed to the network, which can be easily automated by a scene parsing model. Once an artificial scene with desired properties is generated, we then transfer the look of the hallucinated image to the original input image to achieve attribute manipulation in a photorealistic manner.

In Section ~\ref{sec:method}, we present the architectural details of our attribute and layout conditioned scene generation network and the methodologies for effectively training our network. Finally, in Section ~\ref{styletransfer}, we discuss the photo style transfer method that we utilize to transfer the appearance of generated images to the input image.

\begin{figure*}[t]
\centering
\includegraphics[width=\linewidth]{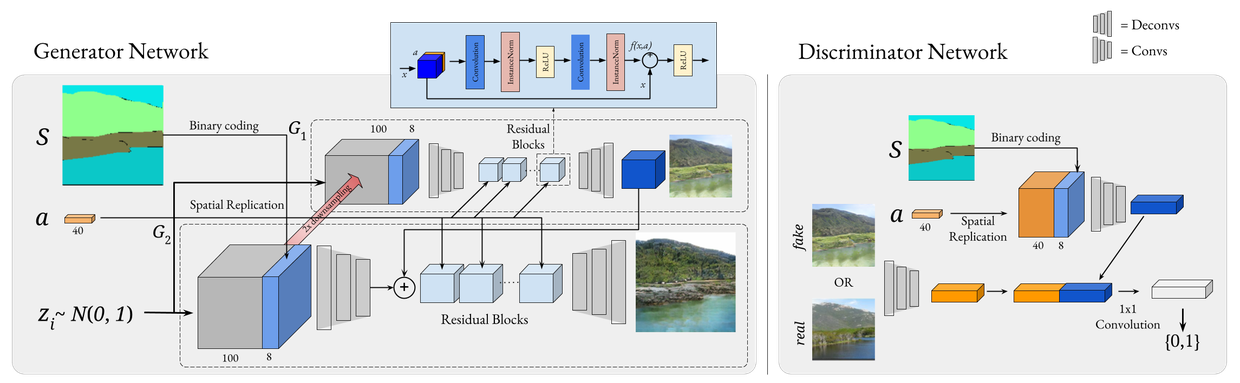}
\vspace{-2mm}
\caption{Our proposed Scene Generation Network (SGN) can generate synthetic outdoor scenes consistent with given layout and transient attributes.}
\label{fig:model}
\end{figure*}

\subsection{Scene Generation}
\label{sec:method}
In this section, we first give a brief technical summary of GANs and conditional GANs (CGANs), which provides the foundation for our scene generation network (SGN). We then present architectural details of our SGN model, followed by the two strategies applied for improving the training process. All the implementation details are included in the Supplementary Material.

\subsubsection{Background.}
\label{sec:GAN}
In Generative Adversarial Networks (GANs) \cite{GPMXWDOCB14}, a discriminator network $D$ and a generator network $G$ play a two-player min-max game where $D$ learns to determine if an image is real or fake and $G$ strives to output as realistic images as possible to fool the discriminator. 
The $G$ and $D$ are trained jointly by performing alternating updates: 
\begin{eqnarray}
\min_G \max_D \mathcal{L}_{GAN}(G,D) & = & E_{x \sim p_{data}(x)}[\log D(x)] + \\\nonumber
&& E_{z \sim p_{z}(z)} [\log \left(1-D(G(z))\right)]
\end{eqnarray}
where $x$ is a natural image drawn from the true data distribution $p_{data}(x)$ and $z$ is a random noise vector sampled from a multivariate Gaussian distribution. The optimal solution to this min-max game is when the distribution $p_G$ converges to $p_{data}$. 

Conditional GANs~\cite{mirza2014conditional} (CGANs) engage additional forms of side information as generation constraints, e.g. class labels~\cite{mirza2014conditional}, image captions~\cite{RAYLSL16}, bounding boxes and object keypoints~\cite{RAMSSL16}. Given a context vector $c$ as side information, the generator $G(z,c)$, taking both the random noise and the side information, tries to synthesize a realistic image that satisfies the condition $c$.
The discriminator, now having real/fake images and context vectors as inputs, aims at not only distinguishing real and fake images but also whether  an image satisfies the paired condition $c$. 
Such characteristics are referred to as match-aware~\cite{RAYLSL16}.
In this way, we expect the generated output of CGAN $x_g$ is controlled by the side information $c$. Particularly, in our model, $c$ is composed of semantic layouts $s$ and transient attributes $a$. 

\subsubsection{Proposed Architecture}
\label{subsec:ascgan}
In our work, we follow a multi-scale strategy similar to that in Pix2pixHD~\cite{wang2018high}. Our scene generator network (SGN), however, takes the transient scene attributes and a noise vector as extra inputs in addition to the semantic layout. While the noise vector provides stochasticity and controls diversity in the generated images, transient attributes let the users have control on the generation process. In more detail, our multi-scale generator network $G=\{G_1, G_2\}$ consists of a coarse-scale ($G_1$) generator and a fine-scale ($G_2$) generator. As illustrated in Fig.~\ref{fig:model}, $G_1$ and $G_2$ have nearly the same architecture, with the exception that they work on different image resolutions. While $G_1$ operates at a resolution of 256 $\times$ 256 pixels, $G_2$ outputs an image with a resolution that is 4$\times$ larger, i.e. 512 $\times$ 512 pixels. Here, the image generated by $G_1$ is fed to to $G_2$ as an additional input in the form of a tensor. In that regard, $G_2$ can be interpreted as a network that performs local enhancements in the fine resolution. 

In our coarse and fine generator networks, while the semantic layout categories are encoded into 8-bit binary codes, transient attributes are represented by a 40-d vector. Input semantic layout map $S$ is of the same resolution with our fine scale image resolution. We concatenate semantic layout $S$ and noise $z$, and feed their concatenation into convolutional layers of $G_1$ and $G_2$ to obtain semantic feature tensors, which are used as input to the subsequent residual blocks. For the coarse scale generator $G_1$, we at first perform a downsampling operation with a factor of $2$ to align the resolutions. Then, spatially replicated attribute vectors $a$ are concatenated to input tensors of each residual block in $G_1$ and $G_2$ to condition the image generation process in regard to input transient scene attributes. Finally, deconvolutional layers are used to upsample the feature tensor of the last residual block to obtain final output images. For fine scale generator $G_2$, semantic feature tensor extracted with the convolutional layers is summed with the feature tensor from the last residual block of coarse generator $G_1$ before feeding into residual blocks of fine scale generator $G_2$. 

The discriminator used in our SGN also adopts a multi-scale approach in that it includes three different discriminators denoted by $D_1, D_2, D_3$ with similar network structures that operate at different image scales. In particular, we create an image pyramid of 3~scales that include real and generated high resolution images, their downsampled versions by a factor of 2 and 4. Our discriminators take tuples of real or synthesized images from different levels of this image pyramid, matching or mismatching semantic layouts and transient attributes and decide whether the images are fake or real, and whether the pairings are valid. That is, the discriminator aims to satisfy
\[
D_k(x_k, a, S) = 
\begin{cases}
1, x_k\in p_{\mathrm{data}} \text{ and }x_k, a, S \text{ correctly match,} \\
0, \text{ otherwise}.
\end{cases}
\]
with $k=\{1,2,3\}$ denoting image scales. Hence, the training our conditional GAN models becomes a multi-task learning problem defined as follows:
\begin{equation}
\min_G \max_{D_1,D_2,D_3} \sum_{k=\{1,2,3\}} \mathcal{L}_{GAN}(G,D_k)
\end{equation}

The architectural details of our Scene Generation Network are given in ~\autoref{tab:detail_arch}. In this table, we follow a naming convention similar to the one used in~\cite{zhu2017unpaired,wang2018high}. For instance, $\mathtt{C_{3}128S_2}$ denotes a Convolution-InstanceNorm-ReLU layer with $128$ filters of kernel size $3\times3$ kernel and stride $2$. $\mathtt{f_{1i}}$ and $\mathtt{f_{2i}}$ represent $i$th internal feature tensors of $G_1$ and $G_2$, respectively. $\mathtt{R512}$ denotes a residual block with filter size $512$ as depicted in Fig.~\ref{fig:model}. Similarly, $\mathtt{D_{3}128S_{0.5}}$ represents a Deconvolution-InstanceNorm-ReLU layer with $128$ filters of kernel size $3\times3$ and stride $0.5$. At the last deconvolution layer $\mathtt{D_{7}3S_1}$, we do not use InstanceNorm and replace ReLU activations with $tanh$. The discriminator resembles a Siamese network~\cite{CHL05,JGLSS13}, where one stream takes the real/generated image as input $x$ and the second one processes the given attributes $a$ and the spatial layout labels $S$ . The responses of these networks are then concatenated $a\Vert S$ and fused via a $1\times 1$ convolution operation.  The combined features are finally sent to fully-connected layers for the binary decision. We use leaky ReLU with slope $0.2$ for our discriminator networks. We do not use InstanceNorm at the input layers. We employ $3$ discriminators at $3$ different spatial scales with $1, 0.5$ and $0.25$ as the scaling factors for both coarse and fine scale generators $G_1$ and $G_2$ during training.

\begin{table}[t]
    \centering
\caption{Architectural details of the generator and discriminator networks.}    
\renewcommand{\arraystretch}{1.2} %
    \begin{tabular}{ccl}
\hline
\multicolumn{3}{c}{\textbf{Generator}}\\
\hline
Net. & Input & Specification\\
\hline
& $z\Vert S$ & $\mathtt{C_{7}64S_1-C_{3}128S_2-C_{3}256S_2-C_{3}512S_2\rightarrow f_{11}}$\\

$G_1$ & $\mathtt{f_{11}}\Vert a$ & $\mathtt{R512-R512-R512-R512-R512\rightarrow f_{12}}$\\

&  $\mathtt{f_{12}}$ & $\mathtt{D_{3}256S_{0.5}-D_{3}128S_{0.5}-D_{3}64S_{0.5}\rightarrow f_{13}}$ \\

& $\mathtt{f_{13}}$ & $\mathtt{D_{7}3S_1}\rightarrow x_{fake}^{256}$\\
\hline

\multirow{3}{*}{$G_2$} & $z\Vert S$ & $\mathtt{C_{7}32S_1-C_{3}64S_2\rightarrow f_{21}}$\\

 & $\mathtt{f_{13} + f_{21}}$ & $\mathtt{R64-R64\rightarrow f_{22}}$\\
 
 & $\mathtt{f_{22}}$ & $\mathtt{D_{3}64S_{0.5}-D_{7}3S_1}\rightarrow x_{fake}^{512}$\\
\hline
\multicolumn{3}{c}{\textbf{Discriminator}}\\
\hline
Net. & Input & Specification\\
\hline
 
  & $x$ & $\mathtt{C_{4}64S_2-C_{4}128S_2-C_{4}256S_2-C_{4}512S_2}\rightarrow \mathtt{f}_x$\\
  
 $D_k$ & $a\Vert S$ & $\mathtt{C_{4}64S_2-C_{4}128S_2-C_{4}256S_2-C_{4}512S_2}\rightarrow \mathtt{f}_c$ \\

  & $\mathtt{f}_x \Vert \mathtt{f}_c$ & $\mathtt{C_{1}512S_1-C_{4}1S_1}\rightarrow [0, 1]$\\
 \hline
    \end{tabular}
    \label{tab:detail_arch}
\end{table}

\subsubsection{Improved Training of SGNs}
\label{subsec:methodology}
Here we elaborate on two complementary training techniques that substantially boost the efficiency of the training process.
\paragraph{Relevant Negative Mining.} 
Training the match-aware discriminator in CGAN resembles learning to rank~\cite{rudin2009margin}, in the sense that a ``real pair''--real image paired with right conditions--should score higher (i.e. classifying into category 1 in this case) than a ``fake pair''--either image is fake or context information is mismatched (i.e. classifying into category 0).
For ranking loss, it has been long acknowledged that naively sampling random negative examples is inferior to more carefully designed negative sampling scheme, such as various versions of hard negative mining~\cite{bucher2016hard,fu2013survey,shrivastava2016training,li2013bootstrapping}. 
Analogously, a better negative mining scheme can be employed by training CGAN, as existing works have been using random sampling~\cite{RAMSSL16}.
To this end, we propose to apply the concept of relevant negative mining~\cite{li2013bootstrapping} (RNM) to sample mismatching layout in training our SGN model. 
Concretely, for each layout $S$, we search for its nearest neighbor $S'$ and set it as the corresponding mismatching negative example for $S$.
In Section~\ref{sec:experiments}, we present empirical qualitative and quantitative results to demonstrate improvement from RNM over random sampling.
We attempted similar augmentation on attributes $a$ by flipping a few of them instead of complete random sampling to obtain the mismatching $a'$ but found such operation hurt the performance, likely due to the flipped attributes being too semantically close to the original ones which cause ambiguity to the discriminator. 

\paragraph{Layout-Invariant Perceptual Loss.} Following the practice of existing works~\cite{chen2017photographic,dosovitskiy2016generating}, we also seek to stabilize adversarial training and enhance generation quality by adding a perceptual loss.
Conventionally, features used for perceptual loss come from a deep CNN, such as VGG~\cite{simonyan2014very}, pretrained on ImageNet for classification task.
However, perceptual loss to match such features would intuitively withhold generation diversity, which opposes our intention of creating stochastic output via a GAN framework. 
Instead, we propose to employ intermediate features trained on outdoor scene parsing with ADE20K. 
The reason for doing so is three-fold: diversity in generation is not suppressed, because scenes with different contents but the same layout ideally produce the same high-level features; the layout of the generation is further enforced thanks to the nature of the scene parsing network; since the scene parsing network is trained on real images, the perceptual loss will impose additional regularization to make the output more photorealistic.
The final version of our proposed perceptual loss is as follows:
\begin{equation}
\mathcal{L}_{percep}(G) = E_{z \sim p_{z}(z); x,S,a \sim p_{data}(S,a)} \left[ \left\| f_P(x)-f_P(G(z,a,S))\right\|^2_2\right],
\end{equation}
where $f_P$ is the CNN encoder for the scene parser network. Our full objective that combines multi-scale GAN loss and layout-invariant feature matching loss thus becomes:
\begin{equation}
\min_G \left( \left(\max_{D=\{D_1, D_2, D_3\}} \sum_{k=1,2,3}\mathcal{L}_{GAN}(G,D_k) \right) +\lambda \mathcal{L}_{percep}(G) \right)
\end{equation}
where $\lambda$ is a scalar controlling the importance of our proposed layout-invariant feature matching loss and is set to 10 in our experiments.
By additionally considering RNM and perceptual loss, we arrive at the training procedure which is outlined in Algorithm~\ref{algorithm}. 

\begin{algorithm}[b]
  \caption{SGN training algorithm}
  \begin{algorithmic}[1]
  	\STATE \textbf{Input:} Training set $\Omega=\{(x,a,S)\}$  with training images $x$, semantic segmentation layouts $S$ and transient attributes $a$.
  	\FORALL{number of iterations}
  	\STATE sample minibatch of paired $x,a,S$
    \STATE sample minibatch of $z_i$ from $\mathcal{N}(0, I)^Z$
    \FORALL{$(x_i,a_i,S_i)$ in $\Omega$}
    \STATE Randomly sample negative $a'_i$ mismatching $x_i$
    \STATE Sample $S'_i$ mismatching $x_i$ via RNM
    \ENDFOR
    \STATE $x_g \leftarrow G(z_i,a_i,S_i)$ \{Forward through generator\}
    
     \FOR {k=1:3}{
        \STATE $\mathcal{L}_{D_k} \leftarrow -(\log{D_k(x, a, S)} + \log{(1-D_k(x_g,a,S)}) + \log{(1-D_k(x,a',S')})$
        \STATE $D_k \leftarrow D_k - \alpha \partial \mathcal{L}_{D_k}/ \partial D_k $ \{Update discriminator $D_k$\}}
    \ENDFOR
    \STATE $\mathcal{L}_G \leftarrow -\log{D(x_g, a,S)} + \lambda \| f_p(x) - f_p(x_g) \|^2_2$
     \STATE $G \leftarrow G - \alpha \partial \mathcal{L}_G/ \partial G $ \{Update generator $G$\}
    \ENDFOR
  \end{algorithmic}
  \label{algorithm}
\end{algorithm} 

\subsection{Style Transfer}
\label{styletransfer}
The main goal in photo style transfer is to successfully transfer visual style (such as color and texture) of a reference image onto another image while preserving semantic structure of the target image. In the past, statistical color transfer methods ~\cite{reinhard2001color, pitie2005n} showed that the success of the style transfer methods highly depend on the semantic similarity of the source and target images. To overcome this obstacle, user interaction, semantic segmentation approaches or image matching methods were utilized to provide semantic relation between source and target images. In addition, researchers explored data driven methods to come up with fully automatic approaches which retrieve the source style image through some additional information such as attributes, features and semantic similarity.

For existing deep learning based photo style transfer methods, it is still crucial that source and reference images have similar semantic layouts to provide successful and realistic style transfer results. Image retrieval based approaches are limited with the dataset and they become infeasible when there is no images with the desired properties. The key distinguishing characteristics of our framework is that we can generate a style image on the fly that has both similar semantic layout with the input image and possess the desired transient attributes, thanks to our proposed SGN model. 
In our framework, for photo style transfer, we consider employing both DPST~\cite{luan2017deep} and FPST~\cite{li2018closed} models.

DPST~\cite{luan2017deep} extends the formalization of the neural style transfer method of~\citet{gatys2016image} by adding a photorealism regularization term that enables the style transfer to be done between same semantic regions instead of the whole image. This property makes DPST very appropriate for our image manipulation system. Although this method in general produces fairly good results,~we observe that it sometimes introduces some smoothing and visual artifacts in the output images, which hurt the photorealism. For that reason, we first apply a cross bilateral filter~\cite{chen2007bilateral} to smooth the DPST's output according to edges in the input image and then apply the post-processing method proposed by~\citet{mechrez2017photorealistic}, which uses screened Poisson equation to make the stylized image more similar to the input image in order to increase its visual quality. 

FPST~\cite{li2018closed} formulates photo style transfer as a two steps procedure. The first step carries out photorealistic image stylization by using a novel network architecture motivated by the whitening and coloring transform~\cite{li2017WCT}, in which the upsampling layers are replaced with unpooling layers. The second step performs a manifold ranking based smoothing operation to eliminate the structural artifacts introduced by the first step. As both of these steps have closed-form solutions, FPST works much faster than DPST. Since FPST involves an inherent smoothing step, in our experiments, we only apply the approach by~\citet{mechrez2017photorealistic} as a post-processing step.

\section{Results and Comparison}
\label{sec:experiments}
 
We first evaluate our scene generation network's ability to synthesize diverse and realistic-looking outdoor scenes, then show attribute manipulation results of our proposed two-stage framework that employs the hallucinated scenes as reference style images. Lastly, we discuss the limitations of the approach. 

\subsection{Attribute and Layout Guided Scene Generation}
\label{ssec:SGN}
Here, we assess the effectiveness of our SGN model on generating outdoor scenes in terms of image quality, condition correctness and diversity. We also demonstrate how the proposed model enables the users to add and subtract scene elements. 

\begin{figure*}[!t]
\centering	
\includegraphics[width=\linewidth, trim=0 0 0 0,clip]{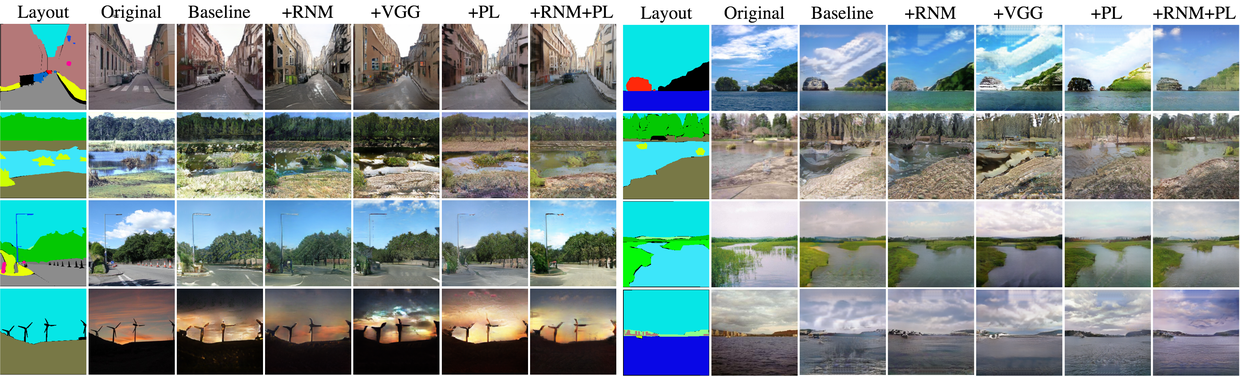}
\caption{Sample scene generation results. In these examples, the input layouts are from the test set, which are unseen during training and the transient attributes are fixed to the original transient attributes. Incorporating Relevant Negative Mining (RNM) and Perceptual Loss (PL) significantly improves the performance of the baseline SGN model in terms of both image quality as well as faithfulness of the end result to conditioned layouts and attributes. Moreover, the way we define our perceptual loss, as compared to commonly used VGG-based one, provides better and more photorealistic results.}
\label{fig:layoutconditioned}
\end{figure*}

\subsubsection{Training Details}
All models were trained with a mini-batch size of $40$ where parameters were initialized from a zero-centered Gaussian distribution with standard deviation of $0.02$. We set the amount of the layout-invariant feature matching loss $\lambda$ to 10. We used the Adam optimizer~\cite{kingma2014adam} with the learning rate value of $2 \times 10^{-4}$ and the momentum value of $0.5$. For data augmentation, we employed horizontal flipping with a probability of 0.5. We trained our coarse-scale networks for $100$ epochs on a NVIDIA Tesla K80 GPU for $3$ days. After training them, we kept their parameters fixed and trained our fine-scale networks for $10$ epochs. Then, in the next $70$ epochs, we updated the parameters of both of our fine and coarse-scale networks together. Our implementation is based on the PyTorch framework. Training of our fine-scale networks took about $10$ days on a single GPU.

\subsubsection{Ablation Study}
\label{sec:conditions}
We illustrate the role of Relevant Negative Mining (RNM) and layout-invariant Perceptual Loss (PL) in improving generation quality with an ablation study. Here we consider the outputs of the coarse-scale generator $G_1$ to evaluate these improvements as it acts like a global image generator. Our input layouts come from the test set, i.e. are unseen during training. Furthermore, we fix the transient attributes to the predictions of the pre-trained deep transient model~\cite{BZGWJ16}.  Fig.~\ref{fig:layoutconditioned} presents synthetic outdoor images generated from layouts depicting different scene categories such as urban, mountain, forest, coast, lake and highway. We make the following observations from these results. 

Attributes of the generated images are mostly in agreement with the original transient attributes. Integrating RNM slightly improves the rendering of attributes but in fact, its main role is to make training more stable. Our proposed layout-invariant PL boosts the final image quality of SGN. The roads, the trees and the clouds are drawn with the right texture; the color distributions of the sky, the water and the field also appear realistic; reasonable physical effects are also observed such as the reflection of the water, fading of the horizon, valid view perspective of urban objects. In our analysis, we also experimented with the VGG-based perceptual loss, commonly employed in many generative models, but as can be seen from Fig.~\ref{fig:layoutconditioned}, our proposed perceptual loss, which performs feature matching over a pretrained segmentation network, gives much better results in terms of photorealism.
Overall, the results with both RNM and PL are visually more pleasing and faithful to the attributes and layouts. 

For quantitative evaluation, we employ the Inception Score (IS)~\cite{salimans2016improved} and the Fr\'{e}chet Inception Distance (FID)~\cite{heusel2017gans}\footnote{In our evaluation, we utilized the official implementations of IS and FID. IS scores are estimated by considering all of the test images from our dataset, which were not seen during training and by using a split size of $10$. While calculating FID scores, we employ all of the test images from our dataset as the reference images.}

The IS correlates well with human judgment of image quality where higher IS indicates better quality. FID has been demonstrated to be more reliable than IS in terms of assessing the realism and variation of the generated samples. Lower FID value means that the distributions of generated images and real images are similar to each other. Table~\ref{tab:scores_alcgan} shows the IS and FID values for our SGN model trained under various settings, together with values for the real image space. These results agree with our qualitative analysis that training with RNM and Perceptual Loss provides samples of the highest quality. Additionally, for each generated image, we also predict its attributes and semantic segmentation map using separately trained attribute predictor by~\citet{BZGWJ16} and the semantic segmentation model by~\citet{zhou2016scene} and we report the average MSE\footnote{The ground truth attributes are scalar values between $0$ and $1$.} and segmentation accuracy again in Table~\ref{tab:scores_alcgan}. Training with the proposed perceptual loss is more effective in reflecting  photorealism and preserving both the desired attributes and the semantic layout better than the VGG-based perceptual loss.

Our SGN model with RNM and Perceptual Loss shows clear superiority to other variants both qualitatively and quantitatively. Thus from now on, if not mentioned otherwise, all of our results are obtained with this model.

\begin{figure*}[!t]
\centering
\includegraphics[width=0.98\linewidth]{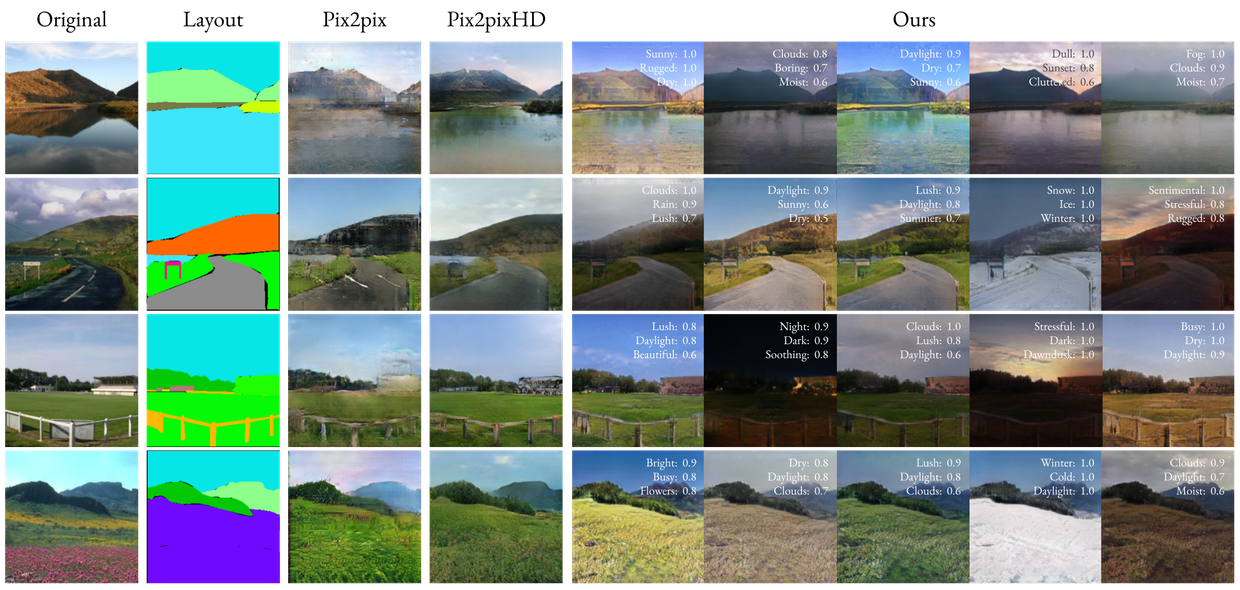}
\vspace{-4mm}
\caption{Comparison of our SGN model against Pix2pix~\cite{isola2016image} and Pix2pixHD~\cite{wang2018high}. Each row shows the original image and the  samples generated according to its corresponding semantic layout. Since our SGN model also takes into account a set of target transient attributes (only the top three most significant ones are shown here for the sake of simplicity), it can generate diverse and more realistic results than the other methods.}
\label{fig:comparison1}
\end{figure*}

\begin{table}[t!]
\centering
\caption{Ablation study. We compare visual quality with respect to Inception Score (IS) and Fr\'{e}chet Inception distance (FID), attribute and semantic layout correctness in terms of average MSE of attribute predictions (Att. MSE) and segmentation accuracy (Seg. Acc.), respectively, via pre-trained models. Our SGN model trained with RNM and PL techniques consistently outperforms the others, including the setting with VGG-based perceptual loss.}
\begin{tabular}{lrrrr}
\hline
{\small{\textbf{Model}}} & {\small{\textbf{IS}}} & {\small{\textbf{FID}}} & {\small{\textbf{Att. MSE}}} & {\small{\textbf{Seg. Acc.}}}\\
\hline
{\small{SGN}} & {\small{3.91}} & {\small{43.77}} &  {\small{0.016}} & {\small{67.70}}\\
$\;\;${\small{+RNM}} & {\small{3.89}} & {\small{41.84}} & {\small{0.016}} & {\small{70.11}}\\
$\;\;${\small{+VGG}} & {\small{3.80}} & {\small{41.87}} & {\small{0.016}} & {\small{67.42}}\\
$\;\;${\small{+PL}} & {\small{4.15}} & {\small{36.42}} & {\small{\textbf{0.015}}} & {\small{70.44}}\\
$\;\;${\small{+RNM+PL}}  & {\small{\textbf{4.19}}} & {\small{\textbf{35.02}}} & {\small{\textbf{0.015}}} & {{\small{\textbf{71.80}}}}\\
{\small{Original}} &  {\small{5.77}} & {\small{0.00}} & {\small{0.010}} & {\small{75.64}}\\
\hline
\end{tabular}
\label{tab:scores_alcgan}
\end{table}

\subsubsection{Comparison with Image-to-Image Translation Models}
We compare our model to Pix2pix~\cite{isola2016image} and Pix2pixHD~\cite{wang2018high} models\footnote{For both of these models, we use the original source codes provided by the authors.}. It is worth mentioning that both of these two approaches generate images only by conditioning on the semantic layout but not transient attributes, and moreover, they do not utilize noise vectors. We provide qualitative comparisons in Fig.~\ref{fig:comparison1}. As these results demonstrate, our model not only generates realistic looking images on par with Pix2pixHD but also has the capability to deliver control over the attributes of the generated scenes. ``Sunset'' attribute makes the horizon slightly more reddish, ``Dry'' attribute increases the brown tones on the trees, ``Snow'' attribute whitens the ground. Also note that the emergence of each attribute tends to highly resonate with part of the image that is most related to the attribute. That is, ``Clouds'' attribute primarily influences the sky, whereas ``Winter'' attribute correlates with the ground, and ``Lush'' tends to impact the trees and the grass. This further highlights our model's reasoning capability about the attributes in producing realistic synthetic scenes.

For quantitative comparison, we compare the IS and FID scores and segmentation accuracy using all $1,338$ testing images in Table ~\ref{tab:pix2pix_comparison} considering both coarse and fine scales. These results suggest that our proposed model produces high fidelity natural images better than Pix2pixHD in both scales. The difference in the segmentation accuracy suggests that Pix2pixHD puts a more strict restraint on the layout whereas our model offers flexibility in achieving a reasonable trade-off between capturing realism in accordance with transient attributes vs. fully agreeing with the layout. Furthermore, in addition to these metrics, we conduct a human evaluation on Figure Eight\footnote{Figure Eight is a web-based data annotation company which can be accessed from \url{https://www.figure-eight.com/}}, asking workers to select among the results of our proposed model and the Pix2pixHD method (for the same semantic layout) which they believe is more realistic. We randomly generate 200 questions, and let 5 different subjects answer each question. We provide the details of our user study in the Supplementary Material. We find that $66\% $ of the subjects picked our results as more realistic. These results suggest that besides the advantages of manipulation over transient attributes, our model also produces higher quality images than the Pix2pixHD model. We also compared our results to the recently proposed Cascaded Refinement Network~\cite{chen2017photographic}, however, it did not give meaningful results on our dataset with complex scenes\footnote{We trained this model using the official code provided by the authors.}.

\begin{table}[!t]
\centering
\caption{Quantitative comparison of layout conditioned image synthesis approaches. Our model consistently outperforms others in both coarse and fine resolutions in terms of photorealism, as measured by IS and FID.}
\begin{tabular}{clrrr}
\hline
& {\small{\textbf{Model}}} & {\small{\textbf{IS}}} & {\small{\textbf{FID}}}  & {\small{\textbf{Seg. Acc.}}}\\
\hline
\parbox[t]{1mm}{\multirow{4}{*}{\rotatebox[origin=lc]{90}{{\small{Coarse$\;\;$}}}}} & {\small{Pix2pix}} & {\small{3.26}} & {\small{76.40}} & {\small{61.93}}\\
& {\small{Pix2pixHD}} & {\small{4.20}} & {\small{47.86}} & {\small{\textbf{75.57}}}\\
& {\small{Ours}}  & {\small{\textbf{4.19}}}& {\small{\textbf{35.02}}} & {\small{71.80}}\\
& {\small{Original}}&  {\small{5.77}} & {\small{0.00}}& {\small{75.64}}\\
\hline
\parbox[t]{1mm}{\multirow{3}{*}{\rotatebox[origin=lc]{90}{{\small{Fine$\,\,\,$}}}}} & {\small{Pix2pixHD}} & {\small{4.87}} & {\small{50.85}} & {\small{\textbf{76.17}}}\\
& {\small{Ours}}  & {\small{\textbf{5.05}}} & {\small{\textbf{36.34}}} & {\small{74.60}}\\
& {\small{Original}}&  {\small{7.37}} & {\small{0.00}}& {\small{77.14}}\\

\hline
\end{tabular}
\label{tab:pix2pix_comparison}
\end{table}

\begin{figure*}[!t]
\centering
\includegraphics[width=0.9\linewidth]{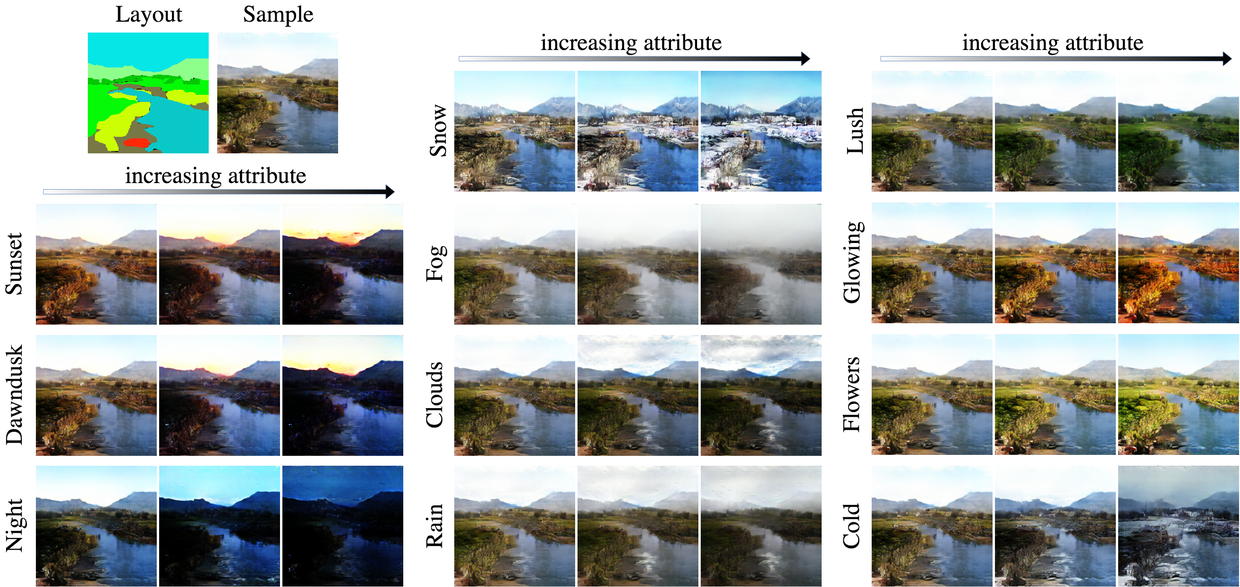}
\caption{Modifying transient attributes in generating outdoor images under different weather and time conditions. Our model's ability of varying with transient attributes contributes to the diversity and photorealism in its generation (more results can be found in the Supplementary Material).}
\label{fig:attcond}
\end{figure*}

\begin{figure}[!t]
\includegraphics[width=\linewidth]{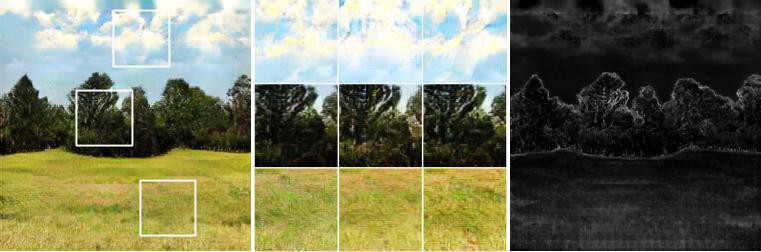}\\
$\;\;$A synthetic scene $\;\;\;$ Stochastic variations $\;\;\;$Standard deviation
\caption{Effect of the noise vector. For an example synthetically generated scene (left), we show close-up views from three different regions (middle) from samples obtained with only changing the random noise. Standard deviation of each pixel over 100 different realizations of the scene (right), which demonstrates that the random noise causes stochastic variations within the irregular or stochastic textural regions.}
\label{fig:noise_effect}
\end{figure}

\subsubsection{Diversity of the Generated Images}
In our framework, a user can control the diversity via three different mechanisms, each playing a different role in the generation process. Perhaps the most important one is the input semantic layout which explicitly specifies the content of the synthesized image, and the other two are the target transient attributes and the noise vector. In Fig.~\ref{fig:attcond}, we show the effect of varying the transient attributes for a sample semantic layout and Fig.~\ref{fig:noise_effect} illustrates the role of noise. If we keep the layout and the attributes fixed, the random noise vector mainly affects the appearance of some local regions, especially the ones involving irregular or stochastic textures such as the sky, the trees or the plain grass. The transient attribute vectors, however, have a more global effect, modifying the image without making any changes to the constituent parts of the scene.

\subsubsection{Adding and Subtracting Scene Elements}
\begin{figure*}[!t]
\centering
\includegraphics[width=1\linewidth]{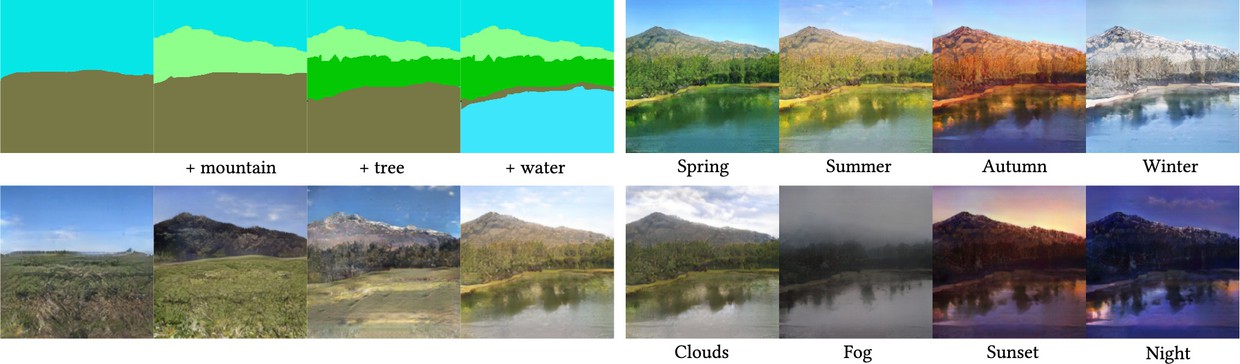}
\vspace{-6mm}
\caption{Gradually adding and removing elements to and from the generated images. We use a coarse spatial layout map (top left) to generate an image from scratch, and then keep adding new scene elements to the map to refine the synthesized images. Moreover, we also show how we can modify the look by conditioning on different transient attributes.}
\label{fig:objectaddremove}
\end{figure*}

Here we envision a potential application of our model as a scene editing tool that can add or subtract scene elements. Fig.~\ref{fig:objectaddremove} demonstrates an example. We begin with a coarse spatial layout which contains two large segments denoting the ``sky'' and the ``ground''. We then gradually add new elements, namely ``mountain'', ``tree'', ``water''. At each step, our  model inserts a new object based on the semantic layout. In fact, such a generation process closely resembles human thought process in imagining and painting novel scenes. The reverse  process,  subtracting elements piece by piece, can be achieved in a similar manner. We sample different random attribute vectors to illustrate how generation diversity can enrich the outcomes of such photo-editing tools and provide a video demo in the Supplementary Material.

\subsection{Attribute Transfer}
\label{ssec:attributetransfer}

\begin{figure*}[!t]
\centering
\includegraphics[width=\linewidth]{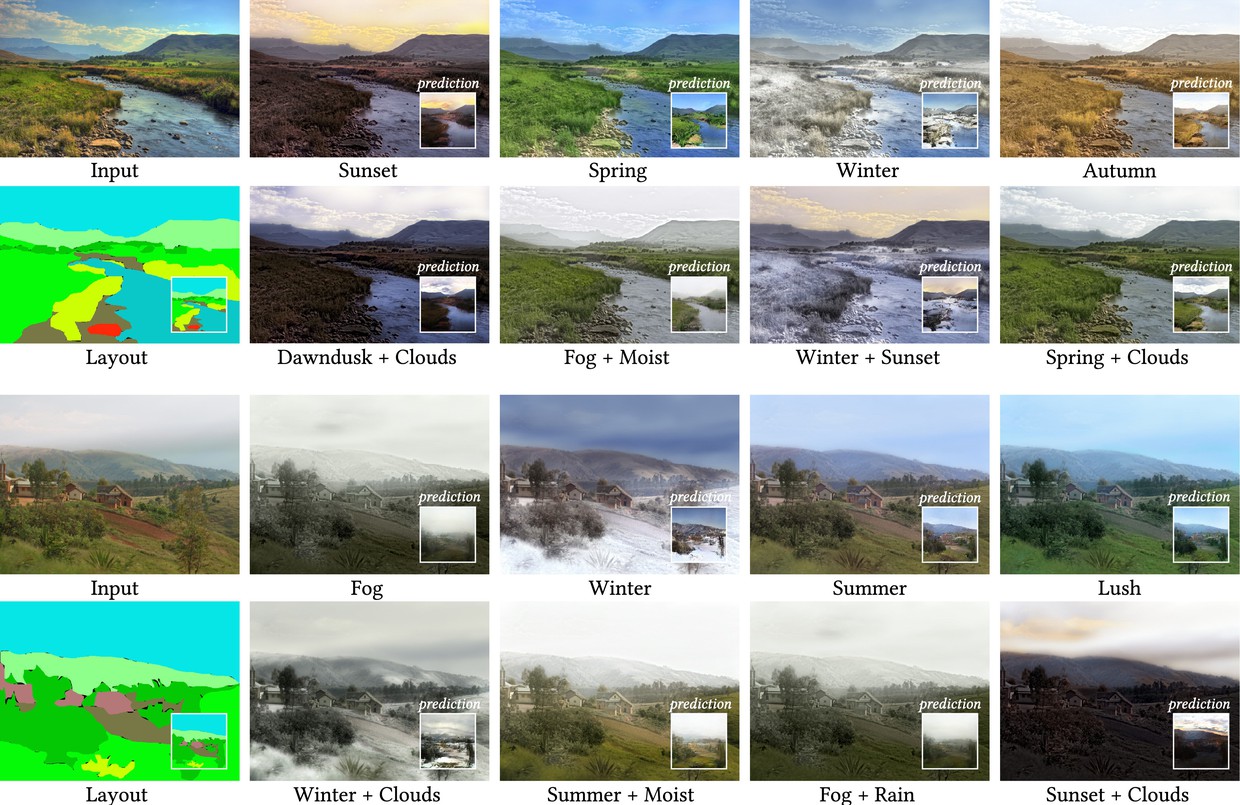}
\vspace{-6mm}
\caption{Sample attribute manipulation results. Given an outdoor scene and its semantic layout, our model produces realistic looking results for modifying various different transient attributes. Moreover, it can perform multimodal editing as well, in which we modify a combination of attributes.}
\label{fig:attribute_transfer}
\end{figure*}

\begin{figure*}[!t]
 \hspace{0.15cm} {\small{Input}} \hspace{1.5cm} {\small{\citet{LRTQH14}}} \hspace{0.15cm} {\small{\citet{LRTQH14} w/ FPST}}  \hspace{0.75cm} Ours\\
\centering
\includegraphics[width=0.725\linewidth]{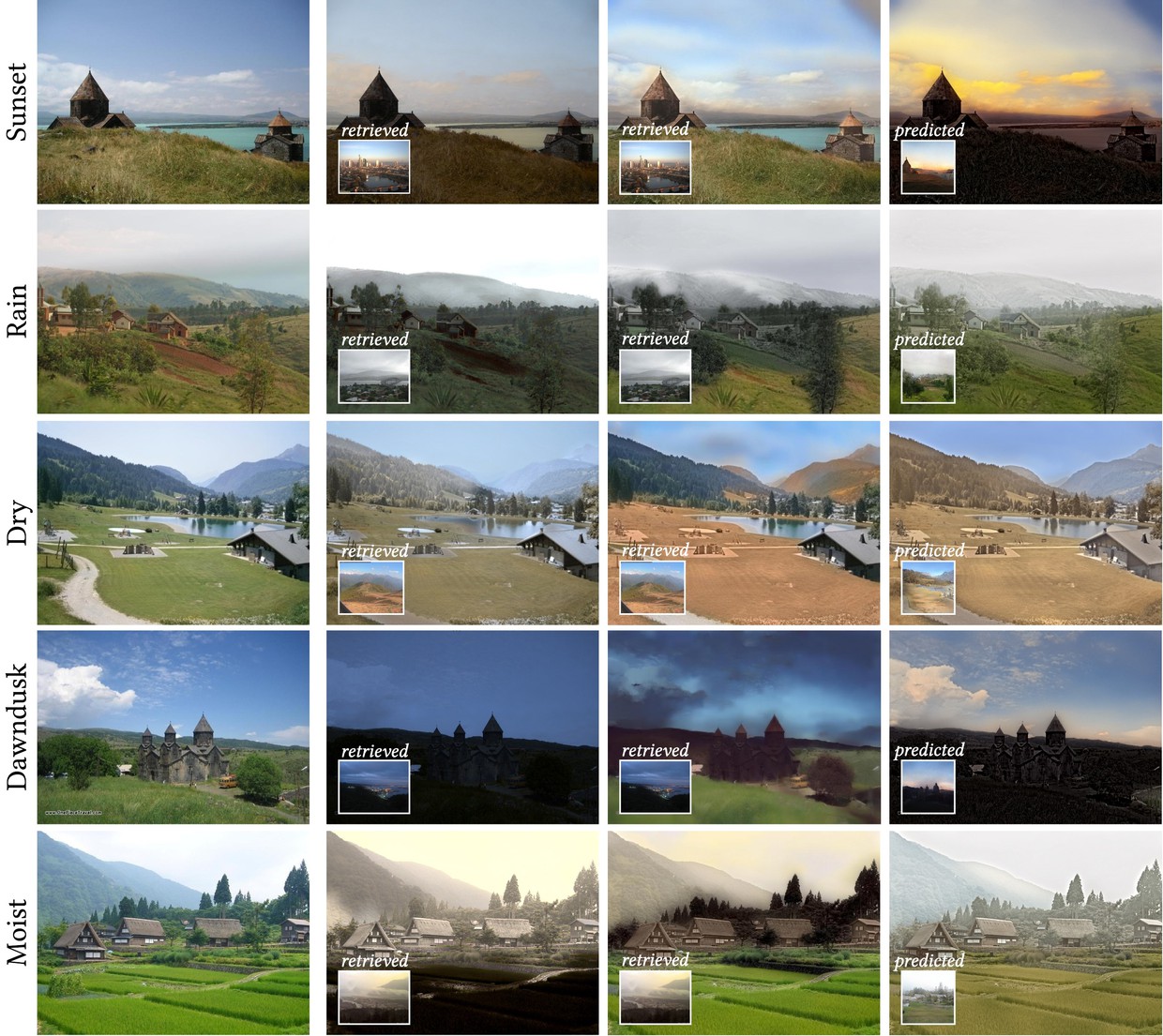}
\caption{Comparison with~\cite{LRTQH14}. In each row, for a given input image (first column), we respectively provide the results of \cite{LRTQH14} using their exemplar-based style transfer method (second column) and FPST~\cite{li2018closed} (third column) between retrieved images and input images, and the results of our method (last column) using FPST~\cite{li2018closed} between generated image by proposed SGN model and input image.}
\label{fig:comparison2}
\end{figure*}

 \begin{figure*}[!t]
\centering
\includegraphics[width=0.955\linewidth]{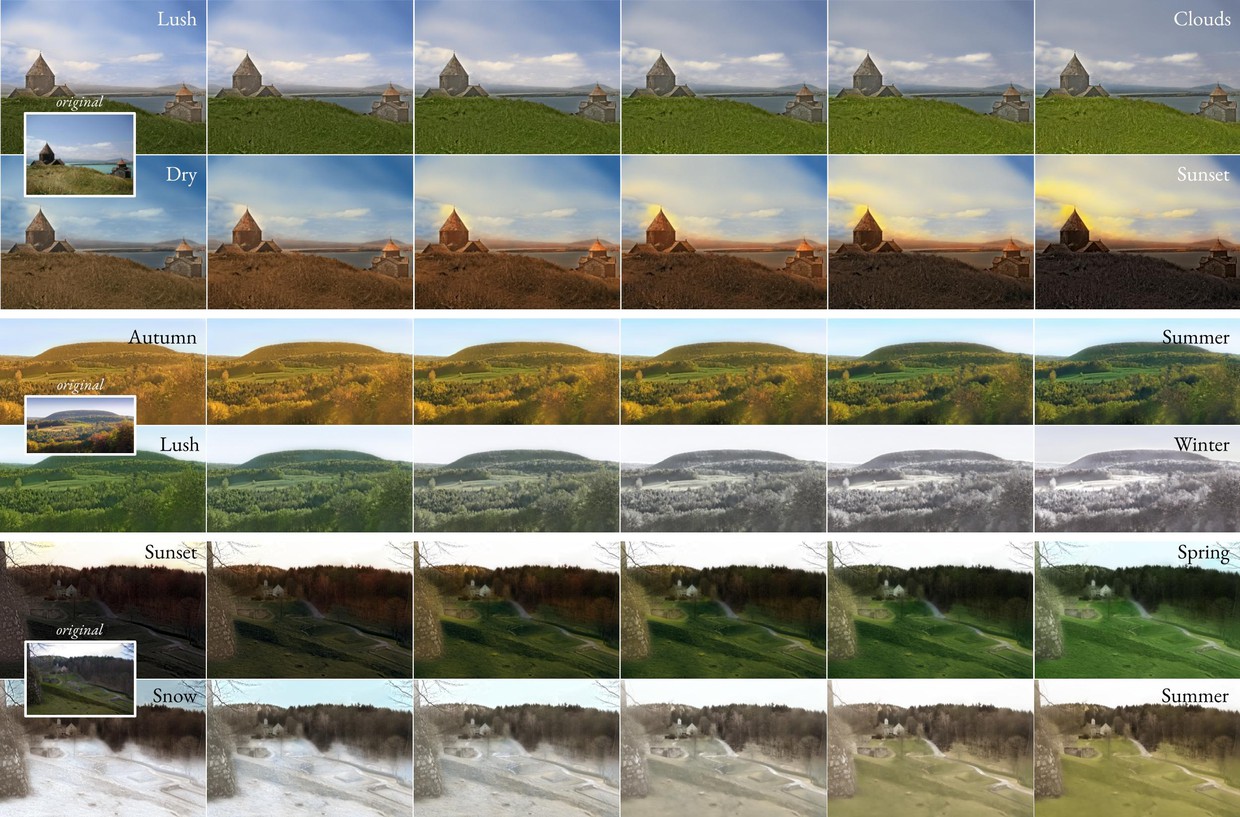}
\vspace{-4mm}
\caption{Our method can produce photorealistic manipulation results for different degrees of transient attributes.}
\label{fig:attributemanifold}
\end{figure*}

We demonstrate our attribute manipulation results in Fig.~\ref{fig:attribute_transfer}. Here we only provide results obtained by using FPST~\cite{li2018closed} as it gives slightly better results in our experiments and also significantly faster than DPST~\cite{luan2017deep}. From now on, unless otherwise stated, all of our attribute transfer results will be the ones obtained with FPST. We provide the results of DPST in the Supplementary Material. As can be seen, our algorithm produces photorealistic manipulation results for many different types of attributes like ``Sunset'', ``Spring'', ``Fog'', ``Snow'', and moreover, a distinctive property of our approach is that it can perform multimodal editing for a combination of transient attributes as well, such as ``Winter and Clouds'' and ``Summer and Moist''. It should be noted that modifying an attribute is inherently coupled with the appearance of certain semantic scene elements. For example, increasing ``Winter'' attribute makes the color of the grass white whereas increasing ``Autumn'' attribute turns them to brown. As another example, ``Clouds'' attribute does not modify the global appearance of the scene but merely the sky region, comparing with ``Fog'' attribute which blurs distant objects; ``Dry'' attribute emphasizes the hot colors, while ``Warm'' attribute has the opposite effect. Some attributes such as ``Fog'', however, have an influence on the global appearance.

In Fig.~\ref{fig:comparison2}, we compare the performance of our method to the data-driven approach of~\citet{LRTQH14}. As mentioned in Section 2, this approach first identifies a scene that is semantically similar to the input image using a database of images with attribute annotations, then it retrieves the version of that scene having the desired properties, and finally, the retrieved image is used as a reference for style transfer. For retrieving the images semantically similar to the source image we also use the Transient Attributes dataset and the retrieval strategy employed by~\citet{LRTQH14}. In fact, since the authors did not publicly share their attribute transfer code, in our experiments, we consider the test cases provided in their project website~\footnote{The test cases we used in our experimental analysis are available at \url{http://transattr.cs.brown.edu/comparisonAppearanceTransfer/testCases.html}.}. In the figure, we both present the reference images generated by our approach and retrieved by the competing method at the right-bottom corner of each output image. For a fair comparison, we also present alternative results of~\cite{LRTQH14} where we replace the original exemplar-based transfer method with FPST~\cite{li2018closed}, which is used in obtaining our results\footnote{Note that, the post-processing method~\citet{mechrez2017photorealistic} is also employed here to improve photorealism.}. As can be seen, our approach produces better results than ~\cite{LRTQH14} in terms of visual quality and as to reflecting the desired transient attributes. These results also demonstrate how style transfer methods are dependent on semantic similarity between the input and style images. Our main advantage over the approach by~\citet{LRTQH14} is that the target image is directly hallucinated from the source image via the proposed SGN model, instead of retrieving the target image from a training set. This makes a difference since the source and the target images always share the same semantic layout. In this regard, our approach provides a more natural way to edit an input image to modify its look under different conditions. 

\begin{table}[t!]
\centering
\caption{User study results for attribute manipulation. The preference rate denotes the percentage of comparisons in which users favor one method over the other.}
\begin{tabular}{lc}
\hline
& {\small{Preference rate}}\\
\hline
{\small{Ours w/ FPST > ~\citet{LRTQH14}}} & {\small{65\%}} \\
{\small{Ours w/ FPST > ~\citet{LRTQH14} w/ FPST}} & {\small{83\%}} \\
{\small{Ours w/ FPST > Ours w/ DPST}} & {\small{52\%}} \\
\hline
\end{tabular}
\label{table:att-user-study}
\end{table}

Additionally, we conducted a user study on Figure Eight to validate our observations. We show the participants an input image and a pair of manipulation results along with a target attribute and force them to select one of the manipulated images which they consider visually more appealing regarding the specified target attribute. The manipulation results are either our results obtained by using DPST or FPST, or those of~\cite{LRTQH14}. We have a total of $60$ questions and we collected at least $3$ user responses per each of these question. We provide the details of our user study in the Supplementary Material. Table~\ref{table:att-user-study} summarizes these evaluation results. We find that the human subjects prefer our approach against the data-driven approach by~\cite{LRTQH14} 65\% of the time. This margin substantially increases when we replace the original exemplar-based transfer part of~\cite{LRTQH14} with FPST as the semantic layouts of retrieved images are most of the time not consistent with those of the input images. We also evaluate the results of our frameworks with FPST and DPST being used as the style transfer network. As can be seen from Table~\ref{table:att-user-study}, the human subjects prefer FPST against DPST but by a very small margin.

The most important advantage of our framework over existing works is that our approach enables users to play with the degree of desired attributes via changing the numerical values of the attribute condition vector. As shown in Fig.~\ref{fig:attributemanifold}, we can increase and decrease the strength of specific attributes and smoothly walk along the learned attribute manifold using the outputs from the proposed SGN model. This is nearly impossible for a retrieval-based editing system since the style images are limited with the richness of the database.

\begin{figure}[!t]
\centering
\includegraphics[width=0.958\linewidth]{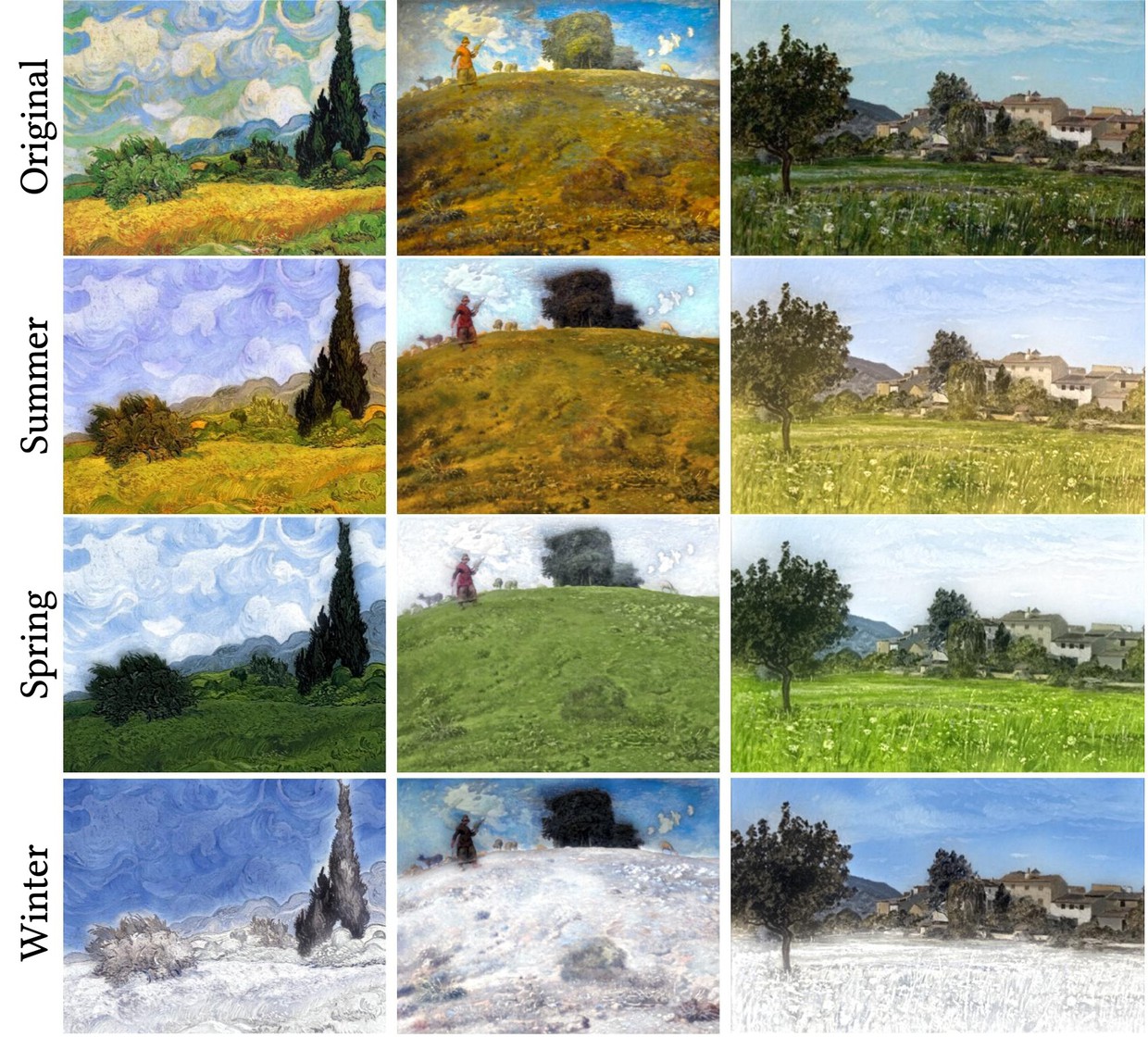}
\caption{Season transfer to paintings. Source images: Wheat Field with Cypresse by Vincent van Gogh (1889), In the Auvergne by Jean-Francois Millet  (1869) and Lourmarin by Paul-Camille Guigou (1868), respectively.}
\label{fig:painting}
\end{figure}

Although our attribute manipulation approach is designed for natural images, we can apply it to oil paintings as well. In Fig.~\ref{fig:painting}, we manipulate transient attributes of three oil paintings to obtain their novel versions depicting these landscapes at different seasons. As can be seen from these results, our model also gives visually pleasing results for these paintings, hallucinating how they might look like if the painters picture the same scene at different times.

\subsubsection{Effect of Post-Processing and Running Times}
\begin{figure}[!t]
\centering
\begin{tabular}{c@{$\;$}c@{$\;$}c}
\includegraphics[width=0.33\linewidth]{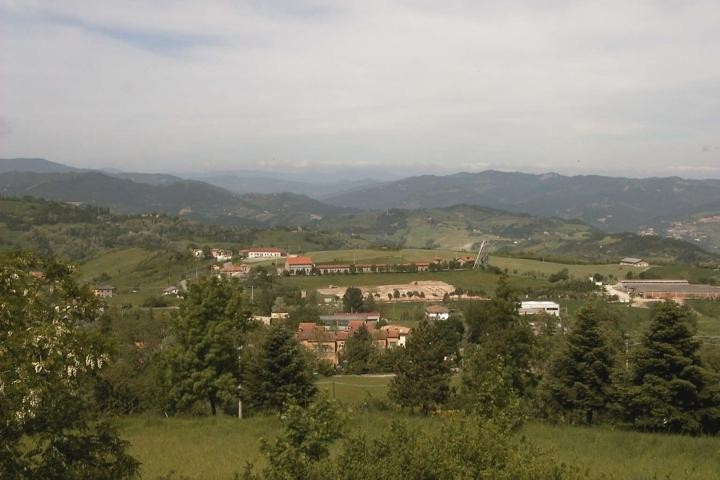} & 
\includegraphics[width=0.33\linewidth]{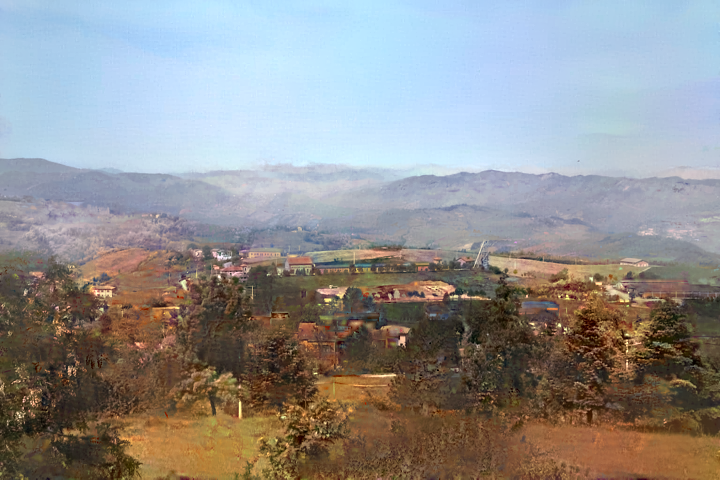} & 
\includegraphics[width=0.33\linewidth]{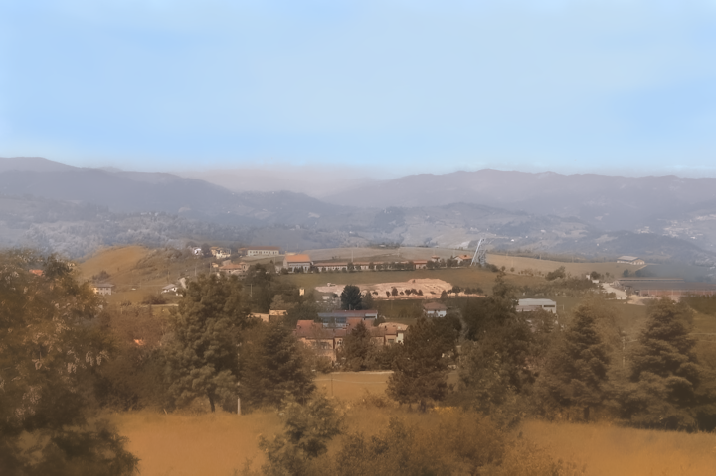} \vspace{-1mm}\\
\small{Input} & \small{DPST} & \small{FPST} \vspace{0.5mm}\\
\includegraphics[width=0.33\linewidth]{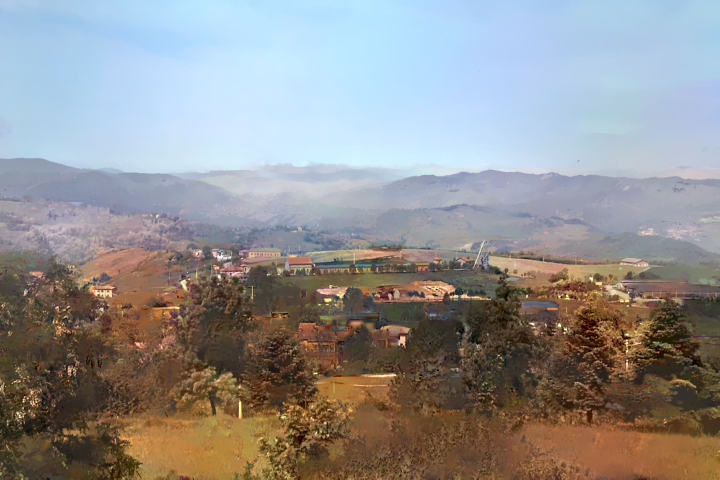} & 
\includegraphics[width=0.33\linewidth]{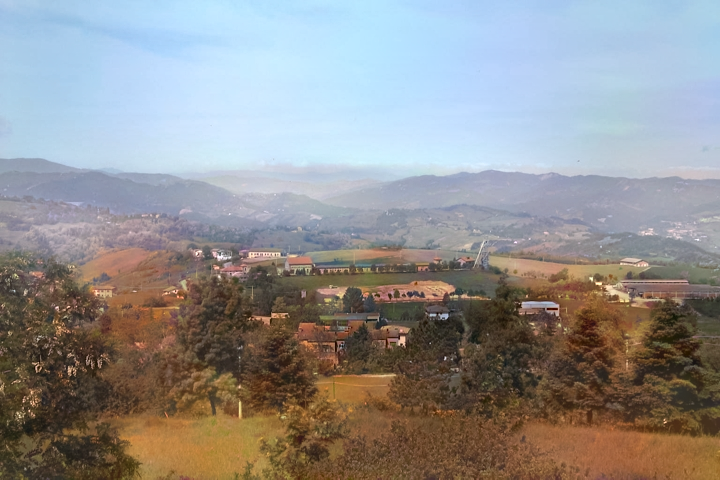} & 
\includegraphics[width=0.33\linewidth]{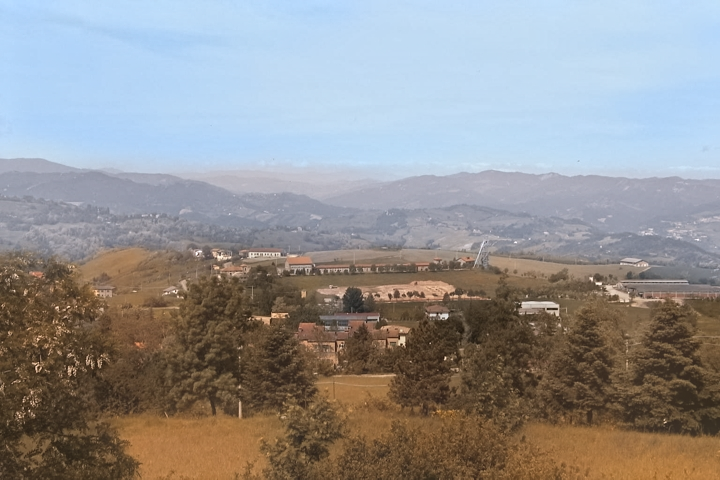} \vspace{-1mm}\\
\small{DPST+BLF} & \small{DPST+BLF+SPE} & \small{FPST+SPE}\\
\end{tabular}
\caption{Effect of post-processing. Top: a sample input image and ``Autumn'' attribute transfer results by our framework with DPST~\cite{luan2017deep} and FPST~\cite{li2018closed}, respectively. Bottom: the impact of various post-processing strategies on the final results. See Section~\ref{styletransfer} for the details.}
\label{fig:post-process}
\end{figure}

We show the effects of the post-processing steps involved in our framework in Fig.~\ref{fig:post-process}. As mentioned in Section~\ref{styletransfer}, for DPST based stylized images, we first apply a cross bilateral filter (BLF)~\cite{chen2007bilateral} and then employ screened Poisson equation (SPE) based photorealism enhancement approach~\cite{mechrez2017photorealistic}. For FPST based stylized images, we only apply SPE as it inherently performs smoothing. As can be seen from these results, the original stylized images demonstrate some texture artifacts and look more like a painting. Our post-processing steps make these stylized images photorealistic and more similar to the given input image. 

\begin{table}[!t]
\centering
\caption{Running time analysis showing the average run time (in seconds) of each component of the proposed model across various image resolutions.}
\renewcommand{\arraystretch}{1.2} 
\begin{tabular}{c@{$\quad$}c@{$\quad\;\;$}c@{$\quad$}c@{$\quad$}c@{$\quad\;\;$}c}
\hline
& {\small{Resolution}} & {\small{SGN}} & {\small{Style Tranfer}} & {\small{Post-Processing}} & {\small{Total}}\\ \hline
\parbox[t]{1mm}{\multirow{3}{*}{\rotatebox[origin=lc]{90}{{\small{DPST}}}}} & {\small{512 $\times$ 256}} & {\small{0.10}} & {\small{1245.31}} & {\small{2.52}} & {\small{1247.93}}\\
& {\small{768 $\times$ 384}} & {\small{0.10}} & {\small{2619.48}} & {\small{4.61}} & {\small{2626.19}}\\
& {\small{1024 $\times$ 512}} & {\small{0.10}} & {\small{4130.27}} & {\small{7.24}} & {\small{4137.51}}\\ \hline
\parbox[t]{1mm}{\multirow{3}{*}{\rotatebox[origin=lc]{90}{{\small{FPST$\;$}}}}} & {\small{512 $\times$ 256}} & {\small{0.10}} & {\small{36.54}} & {\small{1.54}} & {\small{38.18}}\\
& {\small{768 $\times$ 384}} & {\small{0.10}} & {\small{99.34}} & {\small{3.63}} & {\small{103.07}}\\
& {\small{1024 $\times$ 512}} & {\small{0.10}} & {\small{222.20}} & {\small{6.22}} & {\small{228.52}}\\
\hline
\end{tabular}
\label{table:runtime}
\end{table}

In Table~\ref{table:runtime}, we provide the total running time of our framework for manipulating the attributes of an outdoor image. There are three main parts, namely the scene generator network (SGN), the style transfer network, and the post-processing. We report the running time of each of these steps as well. For the style transfer and the post-processing steps, we employ two different versions, one depends on DPST and the other one depends on FPST, and the corresponding smoothing operations. The experiment is conducted on a system with an NVIDIA Tesla K80 graphics card. We consider three different sizes for the input image and report the average run-time for each image resolution. Our FPST-based solution is, in general, much faster than our DPST-based one as most of the computation time is spent on the style transfer step. For images of 1024 $\times$ 512 pixels, while it takes ~4 minutes to manipulate the attributes of an image with FPST, DPST requires ~70 minutes to achieve the task.

\subsection{Effect of Center-cropping}
Our SGN works with a fixed resolution of 512 $\times$ 512 pixels and accepts the semantic layout of the center cropped and resized version of the input image. The style transfer networks consider SGN's output as the target style image and manipulates the input image accordingly. When the image is very wide, like a panorama, center-cropping omits most of the source image. We analyze how this affects the overall performance our framework on a couple of panoramic images from SUN360 dataset~\cite{sun360}. We present attribute manipulation results for one of these images in Fig.~\ref{fig:panorama} and present the rest in the Supplementary Material. We have observed that center-cropping does not pose a serious drawback to our approach, since the style transfer step exploits semantic layouts to constrain color transformations to be carried out between features from the image regions with the same label.

\begin{figure}[!t]
    \centering
    \includegraphics[width=\linewidth]{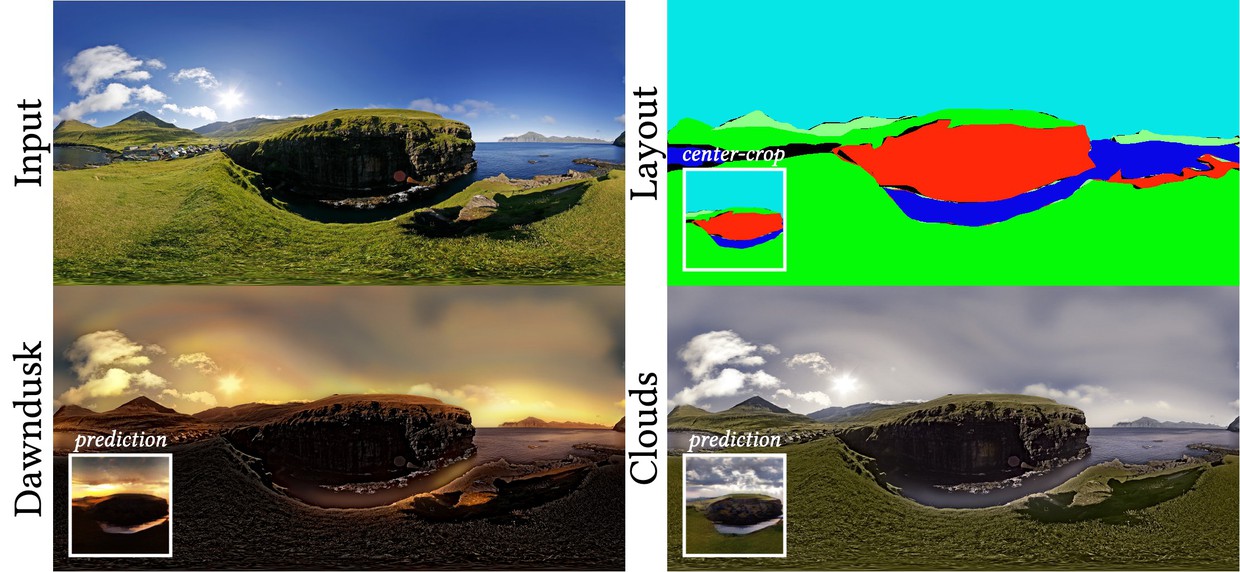}
    \caption{Effect of center-cropping on manipulating a panoramic image.}
    \label{fig:panorama}
\end{figure}

\subsection{Limitations}
\label{ssec:limitations}
\begin{figure}[!t]
\centering
\includegraphics[width=0.95\linewidth]{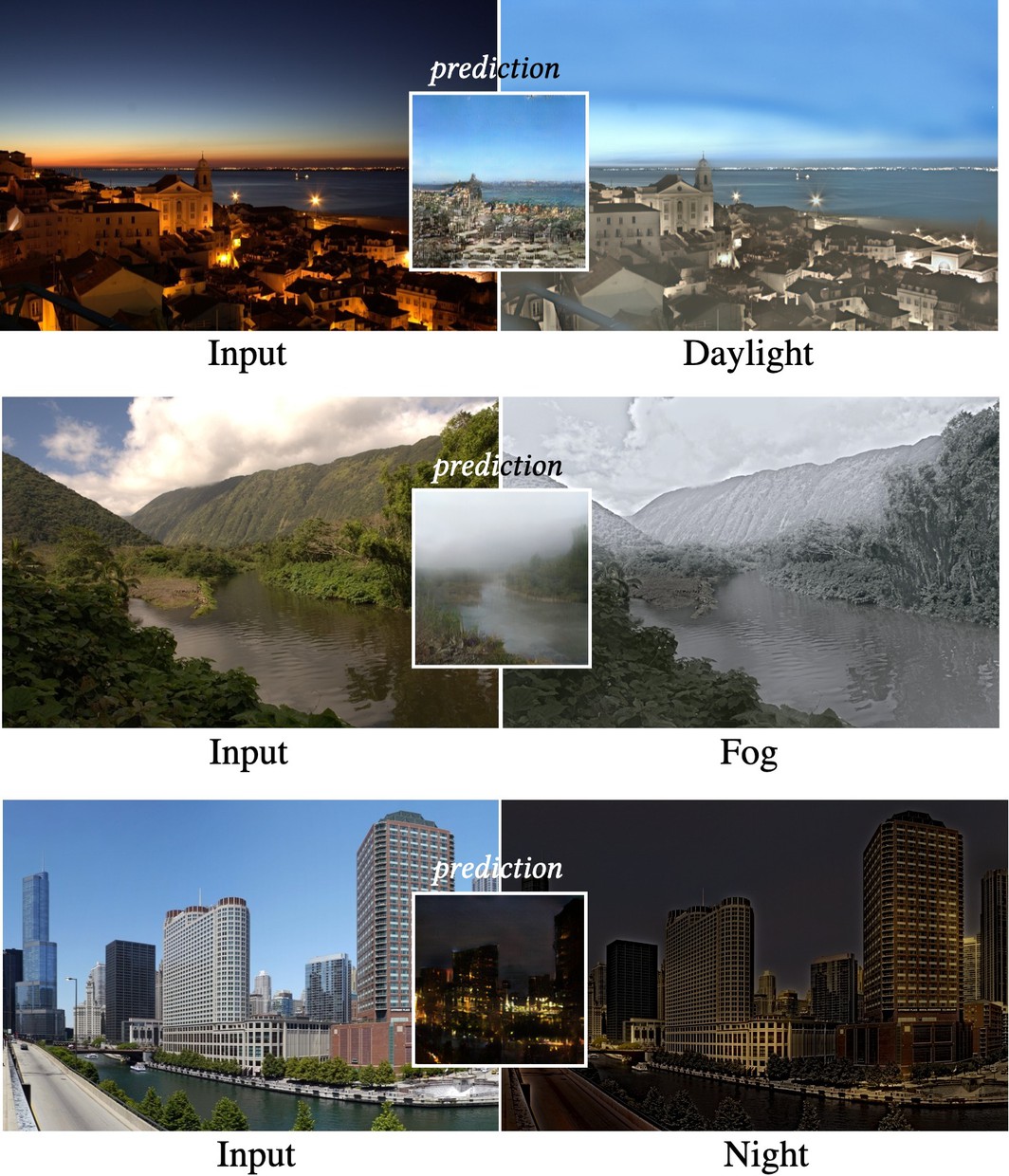}
\vspace{-4mm}
\caption{Example failure cases for our attribute manipulation framework, which are due to the visual quality of synthesized reference style image (top row) and failing of the photo style transfer method (bottom two rows).}
\label{fig:limitation}
\end{figure}

Our framework generally gives quite plausible results, but we should note that it might fail in some circumstances if either one of its components fails to function properly. In Fig.~\ref{fig:limitation}, we demonstrate such example failure cases. In the first row, the photo-realistic quality of the generated scene is not very high as it does note reproduce the houses well. As a consequence, the manipulation result is not very convincing. For the last two scenes, our SGN model hallucinated ``Fog'' and ``Night'' attributes successfully but the style transfer network fails to transfer the looks to the input images.

\section{Conclusion}
\label{sec:conc}
We have presented a high-level image manipulation framework to edit transient attributes of natural outdoor scenes. The main novelty of the paper is to utilize a scene generation network in order to synthesize on the fly the reference style image that is consistent with the semantic layout of the input image and exhibit the desired attributes. Trained on our richly annotated ALS18K dataset, the proposed generative network can hallucinate many different attributes reasonably well and even allows edits with multiple attributes in a unified manner. For future work, we plan to extend our model's functionality to perform local edits based on natural text queries, e.g. add or remove certain scene elements using referring expressions. Another interesting and more challenging research direction is to replace the proposed two-staged model with an architecture that can perform the manipulation in a single shot.

\begin{acks}
This work was supported in part by TUBA GEBIP fellowship awarded to E. Erdem. We would like to thank NVIDIA Corporation for the donation of GPUs used in this research. This work has been partially funded by the DFG-EXC-Nummer 2064/1-Projektnummer 390727645.
\end{acks}

\bibliographystyle{ACM-Reference-Format}
\bibliography{sample-bibliography}